\documentclass{article}
\usepackage[final]{corl_2025} % Uncomment for the camera-ready ``final'' version.
\usepackage{etoolbox} % fix space between abstract header and abstract body
\usepackage[utf8]{inputenc} % allow utf-8 input
\usepackage[T1]{fontenc}    % use 8-bit T1 fonts
\usepackage{fonttable}
\usepackage[english]{babel} %for hyphenation rules
\usepackage[protrusion=true,expansion=true]{microtype}
\usepackage{comment}
\usepackage{cancel}
\usepackage{csquotes}
\usepackage{listings}
\usepackage{nameref}
\usepackage{paralist}
\usepackage{xspace}
\usepackage{setspace}
\usepackage{makeidx}
\usepackage{pdfpages}
\usepackage{times}
\usepackage{cuted} %mixing one column and two column modes
\usepackage{appendix}

\usepackage{amsmath,amssymb} % define this before the line numbering.
\usepackage{amsfonts}       % blackboard math symbols
\usepackage{mathtools}
\usepackage{array} %for table entries to be in center of cell
\usepackage{nicefrac}       % compact symbols for 1/2, etc.
\usepackage{breqn}          % dmath multiline
\usepackage{textcomp}
\usepackage{bm} %bold symbols in math mode
\usepackage{pifont}
\usepackage[
  separate-uncertainty = true,
  multi-part-units = repeat
]{siunitx}
\usepackage{mathrsfs}

\usepackage{graphicx}
\usepackage{wrapfig}
\usepackage{adjustbox} %add additional box operations in addition to resizebox
\usepackage{epsfig}
\usepackage{float}
\usepackage{subcaption}
\usepackage[font=small,labelfont=bf]{caption}
\usepackage{lscape}      %Useful for wide tables or figures.

\usepackage{tikz}  %make graphs
\usepackage{makecell}
\usepackage{overpic}
\newsavebox{\subfigbox}

\usepackage{algorithm}
\usepackage[noend]{algpseudocode}

\usepackage{array} %for table entries to be in center of cell
\usepackage{tabularx}
\usepackage{multirow}
\usepackage{multicol}
\usepackage{booktabs}   % Publication quality tables
\usepackage{tablefootnote}

\usepackage{paralist}
\usepackage{enumitem}
\setlist[itemize]{noitemsep,leftmargin=*,topsep=0in}
\setlist[enumerate]{noitemsep,leftmargin=*,topsep=0in}

\usepackage{url}            % simple URL typesetting
\PassOptionsToPackage{table}{xcolor}
\usepackage{color,colortbl} %superceded by xcolor
\usepackage{soul}

\PassOptionsToPackage{capitalize, noabbrev}{cleveref}

\hypersetup{pagebackref=false,breaklinks=true,colorlinks,urlcolor=blue,citecolor=blue,linkcolor=blue,bookmarks=false}
\usepackage{blindtext}
\usepackage{bbm}

\usepackage{lipsum}

\usepackage{titlesec}
\titlespacing{\section}{0pt}{0.3\baselineskip}{0.25\baselineskip}
\titlespacing{\subsection}{0pt}{0.2\baselineskip}{0.15\baselineskip}
\titlespacing{\subsubsection}{0pt}{0.05\baselineskip}{0.03\baselineskip}

\renewcommand{\paragraph}[1]{\vspace{0.2em}\noindent\textit{#1} --}
\definecolor{color1}{rgb}{.6,.4,.05}
\definecolor{color2}{rgb}{0,.7,.7}
\definecolor{color3}{rgb}{0.35,0.75,0.0}
\definecolor{color4}{rgb}{0.4,0.8,0}
\definecolor{color5}{rgb}{0.5,0.0,0.5}
\definecolor{revision_color}{rgb}{1,0,0}

\usepackage{titlesec}
\titlespacing{\section}{0pt}{0.3\baselineskip}{0.25\baselineskip}
\titlespacing{\subsection}{0pt}{0.2\baselineskip}{0.15\baselineskip}
\titlespacing{\subsubsection}{0pt}{0.05\baselineskip}{0.03\baselineskip}

\renewcommand{\paragraph}[1]{\vspace{0.2em}\noindent\textbf{#1}:}
\newcommand{\ourmethod}{AnyPlace\xspace}
\newcommand{\numtasks}{16\xspace}
\newcommand{\energyvlm}{AnyPlace-EBM\xspace}

\newrobustcmd\B{\DeclareFontSeriesDefault[rm]{bf}{b}\bfseries}
\newcommand{\appendixref}[1]{\hyperref[#1]{Appendix \ref{#1}}}

\title{ AnyPlace: Learning Generalizable Object Placement for Robot Manipulation}

\author{
\textbf{Yuchi Zhao}\textsuperscript{1, 2}, 
\textbf{Miroslav Bogdanovic}\textsuperscript{1, 2}, 
\textbf{Chengyuan Luo}\textsuperscript{3}, 
\textbf{Steven Tohme}\textsuperscript{4},\\
\textbf{Kourosh Darvish}\textsuperscript{1, 5}, 
\textbf{Al\'{a}n Aspuru-Guzik}\textsuperscript{1,2,5}, 
\textbf{Florian Shkurti}\textsuperscript{1,2}, 
\textbf{Animesh Garg}\textsuperscript{6}\\
\textsuperscript{1}University of Toronto,
\textsuperscript{2}Vector Institute,
\textsuperscript{3}Shanghai Jiao Tong University,\\
\textsuperscript{4}Wilfrid Laurier University,
\textsuperscript{5}Acceleration Consortium,
\textsuperscript{6}Georgia Institute of Technology
}

\begin{document}
\maketitle

%===============================================================================

\vspace{-5mm}
\begin{figure}[h]
\centering
\includegraphics[width=0.95\columnwidth]{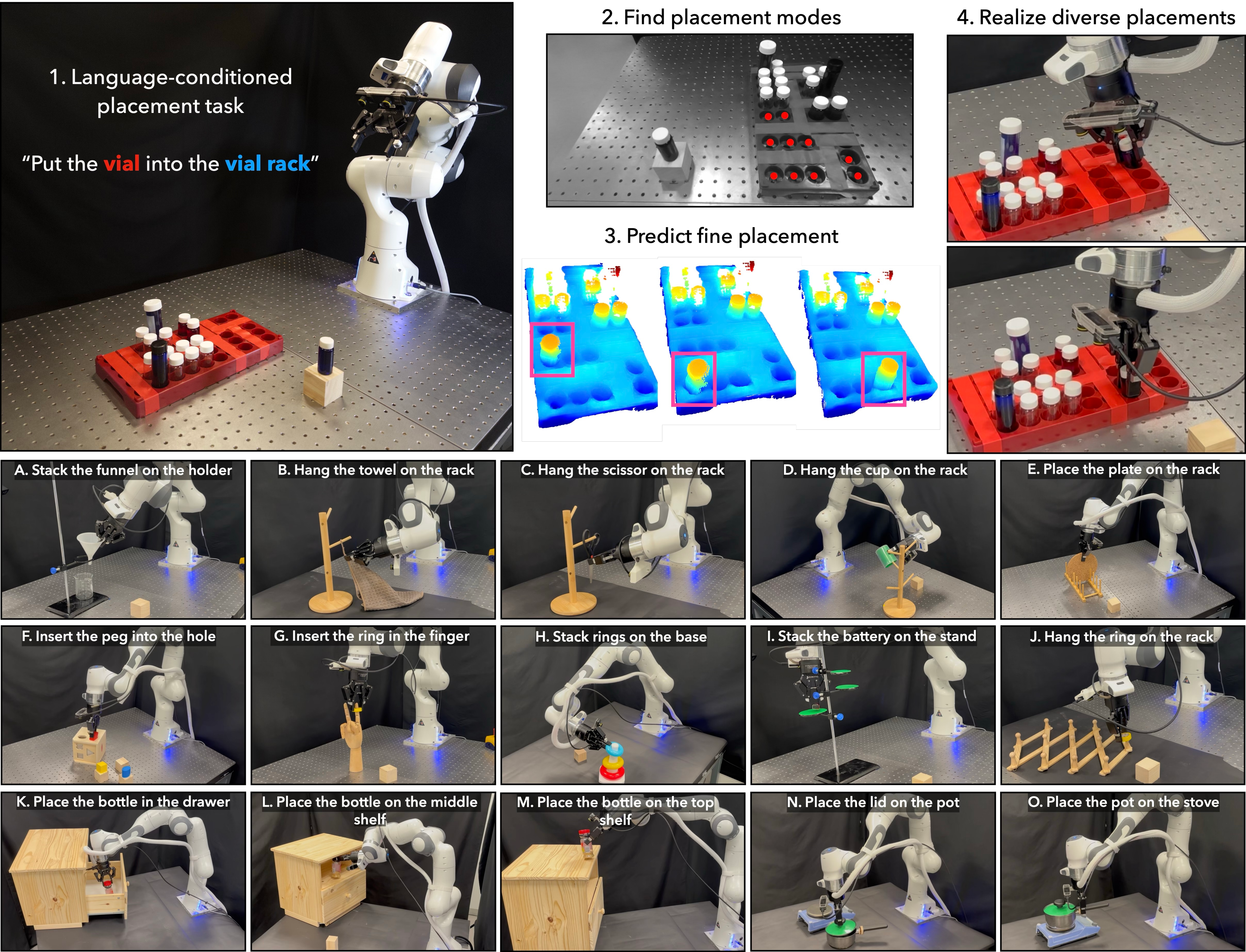}
\caption{\textbf{Execution of the \ourmethod approach by the robot.} (1) Given a language description of a placement task, the robot first captures an RGBD image of the scene using its eye-in-hand camera. (2) A segmentation model and a VLM are used to segment objects and suggest possible placement locations. (3) Multiple placement poses are predicted for objects around suggested locations. (4) The robot realizes placement into any of the predicted poses. (A-O)
%We evaluate AnyPlace on 14 different placement tasks using a real robot of varying precision requirement.
\ourmethod shows generalization and robustness in predicting placement poses across \numtasks tasks in the real world, despite being trained purely on a small synthetic dataset.}

 % \hl{ag: I count 16 here in image?! and there are newer tasks in appendix 15+}

\label{fig:teaser}
\end{figure}
\vspace{-3mm}

\begin{abstract}
Object placement in robotic tasks is inherently challenging due to the diversity of object geometries and placement configurations.
We address this with AnyPlace, a two-stage method trained entirely on synthetic data, capable of predicting a wide range of feasible placement poses for real-world tasks.
Our key insight is that by leveraging a Vision-Language Model (VLM) to identify approximate placement locations, we can focus only on the relevant regions for precise local placement, which enables us to train the low-level placement-pose-prediction model to capture multimodal placements efficiently.
For training, we generate a fully synthetic dataset comprising 13 categories of randomly generated objects in 5370 different placement poses across three configurations (insertion, stacking, hanging) and train local placement-prediction models.
We extensively evaluate our method in high-fidelity simulation and show that it consistently outperforms baseline approaches across all three tasks in terms of success rate, coverage of placement modes, and precision.
In real-world experiments, our method achieves an average success and coverage rate of 76\% across three tasks, where most baseline methods fail completely.
We further validate the generalization of our approach on \numtasks real-world placement tasks, demonstrating that models trained purely on synthetic data can be directly transferred to the real world in a zero-shot setting. 
% \hl{14, ag: inconsistent}
%In real-world experiments, we quantitatively shows that our method achieves an average success/coverage rate of 76\% on three tasks where most of baseline methods fail tasks. WE also conduct 14 different placement tasks and show how our approach directly transfers models trained purely on synthetic data to the real world, where it successfully performs placements in scenarios where other models struggle --- such as with varying object geometries, diverse placement modes, and achieving high precision for fine placement.  
More at: \href{https://any-place.github.io}{\textcolor{blue}{any-place.github.io}}.

% \miroslav{Maybe use em dash directly — instead of --- when copying into open review, although I think it can support latex}
\end{abstract}    

\section{Introduction}

Placing objects is a fundamental task that humans perform effortlessly in daily life, from setting items on a table to inserting cables into sockets. On the other hand, enabling a robot to perform such tasks can often be highly challenging. The challenges arise from the various constraints of different placement tasks and the difficulty of generalizing to unseen objects. Additionally, predicting multimodal placement outputs, which encompass a range of valid locations and modes, remains challenging, particularly when multiple feasible solutions exist. Existing methods are often task-specific, using \textit{a large number of demonstrations for a single placement task}, such as hanging objects on racks \cite{simeonov2023rpdiff}, hoping that the robot can generalize to unseen objects. Alternatively, few-shot approaches \cite{simeonov2021neuraldescriptorfieldsse3equivariant, 10160423, pmlr-v205-simeonov23a, pan2022taxpose, ryu2023diffusionedfsbiequivariantdenoisinggenerative, ryu2023equivariantdescriptorfieldsse3equivariant, gao2024riemann, eisner2024deep, huang2024imagination, huang2024matchpolicysimplepipeline} focus on learning object placement with a few demonstrations, aiming for the model to \textit{replicate the same placement operation} across random initial configurations of similar objects and setups. However, both struggle with generalization and scalability.

We reframe the placement problem as a \textit{pairwise shape mating}~\cite{chen2022neural},  allowing us to \textit{unify multiple placement tasks into a single learning objective}. 
In this work, we address generalizable object placement that is robust to different objects and capable of predicting diverse and precise placement poses across various tasks, such as placing in open areas, inserting in fine slots, and hanging. 
We have developed a fully synthetic dataset that captures three common placement configurations: \textit{inserting, stacking, and hanging}. Furthermore, we develop a placement prediction algorithm that consists of a high-level placement position proposal module and a low-level placement pose prediction model. We use a spatial VLM to propose all possible placement locations and extract the local point cloud regions based on them. By providing a coarse region for the low-level module to focus on, the model can effectively generalize across objects, learn their geometry, and capture diverse placement configurations. We use a diffusion model for fine pose prediction, conditioned on a small local point cloud region, which enables precise and multimodal placement pose predictions. We demonstrate the effectiveness of our methods across a range of placement tasks in simulation, as well as diverse real-world tasks, outperforming the baseline models in terms of success rate and coverage in both cases.
The key contributions of our work are: 
\begin{enumerate}
    \vspace{-1mm}
  \item We propose a novel object placement approach that leverages a VLM to reason about potential placement locations and a low-level pose prediction model to predict placement poses based solely on the region of interest. We show that this coarse-to-fine mechanism allows us to significantly improve performance with respect to baseline methods in terms of success rate, precision, and placement mode coverage.
    \item We develop a data generation pipeline and build a fully synthetic dataset containing thousands of generated objects and capturing a wide range of local placement configurations. 
  \item We demonstrate our approach generalizes to the real world on 16 placement tasks with novel objects and varying degrees of precision requirements.
  \vspace{-1mm}
\end{enumerate}

\section{Related Work}

The problem of robot pick-and-place is typically formulated in two ways: object rearrangement and direct end-effector pose prediction.

\textbf{Object Rearrangement}
In object rearrangement, the goal is to train a model to predict the relative transformation of the object from its initial pose to its final placement pose. %Assuming the grasping pose is generated by existing models, the final placing pose of the end-effector is calculated by multiplying the predicted relative transformation with the grasping pose. 
In this setting, many of the works focus on predicting explicit task-relevant features of both objects and then solving for the relative pose through optimization or regression. Specifically, the Neural Descriptor Fields (NDF) series of papers \cite{simeonov2021neuraldescriptorfieldsse3equivariant, 10160423, pmlr-v205-simeonov23a} learn the occupancy field of point clouds as a representation. 
A fixed set of keypoints on the placement object queries features from the target’s field, and the best transformation is estimated by matching these to demonstration features. %Then, a set of predefined keypoints attached to the placement object is used to interact with and query the occupancy feature field of the target object. By matching the queried features at each point to features collected during demonstrations, the optimal object transformation is determined. 
TaxPose \cite{pan2022taxpose} leverages transformer-based cross-attention to predict corresponding points between two objects and uses differentiable singular value decomposition (SVD) to solve for the relative transformation. To guarantee the placement pose prediction model is robust to SE(3) transformations, i.e., SE(3)-equivariant, methods \cite{ryu2023diffusionedfsbiequivariantdenoisinggenerative, ryu2023equivariantdescriptorfieldsse3equivariant, gao2024riemann} explicitly predict per point type-0 and type-1 features for object point clouds and then solve an optimization problem to align these features into specific configurations based on demonstrations. All of these methods operate in a few-shot setting and can predict a single placement pose given two objects. It is crucial for models to capture and predict a distribution of placement poses, as not every placement pose is realizable by a robot due to its kinematic constraints.  RPDiff \cite{simeonov2023rpdiff}, by contrast, trains a transformer with a diffusion mechanism on a large dataset, gradually denoising the object placement pose. However, their experiments reveal that the coverage of possible placement locations is incomplete. The fixed-size cropping mechanism used during diffusion may also struggle to generalize to objects of varying sizes. Additionally, a recent study \cite{ding2024opendor} samples multiple stable placements in a simulation and employs a VLM to select the appropriate mode based on a language query. While these modes are discrete, each mode allows for the rotation of objects along their axis of symmetry, resulting in valid placement poses that form a continuous distribution. 

\textbf{Direct Pick and Place End-effector Pose Prediction} An alternative approach to the pick-and-place task is predicting the robot's end-effector pose directly. M2T2 \cite{yuan2023m2t2} and Pick2Place \cite{10160736} focus on planar object placement in cluttered scenes. M2T2 \cite{yuan2023m2t2} employs a multi-task transformer with separate decoders to predict grasp poses and placement location affordance maps for each discrete bin of rotation. Other works, like Pick2Place \cite{10160736}, concentrate on predicting key end-effector poses to accomplish specific tasks. RVT \cite{goyal2023rvt} and RVT-2 \cite{goyal2024rvt} also utilize a transformer, leveraging multiview RGB images of the scene to predict heatmaps for the robot's next end-effector location. Coarse-to-fine Q-attention \cite{james2022coarsetofineqattentionefficientlearning}, on the other hand, leverages the scene's voxel to identify the most interesting spatial point at the current resolution. This point becomes the voxel centroid for the next refinement step, enabling the model to gather more accurate 3D information. Another line of work on contemporary VLMs and vision foundation models (VFMs) \cite{Chen_2024_CVPR} finds that they often lack reliable spatial reasoning abilities, limiting their effectiveness in fine-grained manipulation tasks such as peg-in-hole insertion, where precise object placement is critical.

% \subsection{Multimodal prediction in robot manipulation}
% Consider a mug that can be hung on a rack in various poses at different pegs. Multimodality of placement poses is common in manpiulation. Recently advancements in behavior cloning methods, such as IBC energy-based model \cite{florence2021implicit} and Diffusion Policy \cite{chi2023diffusionpolicy}, have attempted to capture this multimodality. For diffusion policy specifically, it learns to denoise action trajectories starting from a random noise distribution, enabling the model to capture the underlying distribution of possible solutions. Another approach to modeling these distributions is through Variational Autoencoders (VAEs). BeT \cite{shafiullah2022behaviortransformerscloningk} and VQ-Bet \cite{lee2024behavior}, for instance, trained a VQ-VAE \cite{oord2018neuraldiscreterepresentationlearning} to discretize and encode continuous actions into latent representations, making it easier to handle multimodal and high-dimensional behavior data. To train the entire policy based on image input, they employed a transformer-based model alongside the VAE decoder to predict sequences of actions. The authors demonstrated that their approach outperformed the Diffusion Policy \cite{chi2023diffusionpolicy} on common benchmarks.

\section{\ourmethod: Generalizable Object Placement}
\label{sec:method}

To enable a robot to execute diverse object placements in a scene, we propose decomposing the placement pose prediction problem into two stages: a high-level coarse placement location proposal stage and a low-level fine placement pose prediction stage. For the high-level task, we incorporate a vision-language model, trained to output 2D keypoint locations in an image based on a given text prompt.
% Subsequently, we crop the region of interest from the scene’s point cloud, centering it around the proposed keypoints.
A small local region around the candidate placement location can then be extracted for the low-level pose-prediction model.
This simplifies the low-level pose prediction problem significantly by using a much smaller point cloud as input and improves generalization overall, as \textit{features outside the local region do not influence the prediction}.
% This allows the low-level pose-prediction to focus exclusively on predicting placement poses based on the local region relevant to the task, significantly simplifying the problem.
This allows us to focus on a limited set of general placement types and utilize a fully synthetic dataset, but have the final model be effective in a broad range of real-world placement tasks. Additionally, the high-level prediction stage enables the identification of multiple placement modes.
% By doing so, we can focus on a limited set of placement types while enabling the model to generalize to a wide range of real-world scenarios. To support the low-level module in learning diverse placement poses, we create a synthetic dataset that incorporates a broad range of object geometries and placement configurations.

% \mb{Maybe to add here explicitly: With current methods it is difficult to have both fine precision and coverage of possibilities in one model}

% \mb{
% Give context of the idea: Get very local region. Now the placement model can be trained in a more general way and be more precise (not just precise, different modes,...)

% We train the local model using synthetic data. We highlight the data generation process, model architecture, and training in \cref{subsec:placement_model}.

% We utilize a VLM-based pipeline to acquire required language-conditioned local regions at test time. We give details of the pipeline in \cref{subsec:detection_pipeline}.

% We rely on a grasping model and a motion planner with collision avoidance for realizing the actual placement. We cover elements of that module in \cref{subsec:placement_pipeline}.
% }

% \subsection{Problem setup}
\noindent \textbf{Problem setup.}
We formulate the object placement task as predicting relative transformations. Specifically, given an input tuple \(\{D, I\}\), where \( D \) represents the language description of the placement task and \( I \) is an RGBD image of the scene, our goal is to predict a set of rigid transformations \( \{T_n\}_{n=1}^{N} \subset \mathrm{SE}(3) \) that move the target object \( C \) from its current position to all viable placement locations on the base object \( B \) that satisfy language conditioning \( D \).  Assuming the grasping poses \( T_{\text{grasp}} \) are provided by a grasp prediction model, the final end-effector pose \( T_{\text{place}} \) can be computed using the predicted relative transformation between the initial object pose and its final placement pose as $T_{\text{place}} = T_n T_{\text{pick}}.$

% \mb{we want to predict a set of N rigid transformations moving the desired object from current position to all viable placement poses that satisfy conditioning given in task \( I \):
% \( T_{n}\)

% product t = 1 to t_max T_{n}^(t)
% In low level: \( T_{n} = \text{product} \)
% }

% \mb{I think maybe VLM $\rightarrow$ dataset $\rightarrow$ then low-level. Nicer story: VLM gets us local regions, that's what we need in the dataset and then we talk about using it right after}

\begin{figure}[t]
    \setlength{\belowcaptionskip}{1pt} 
    \centering
    \includegraphics[width=0.94\textwidth]{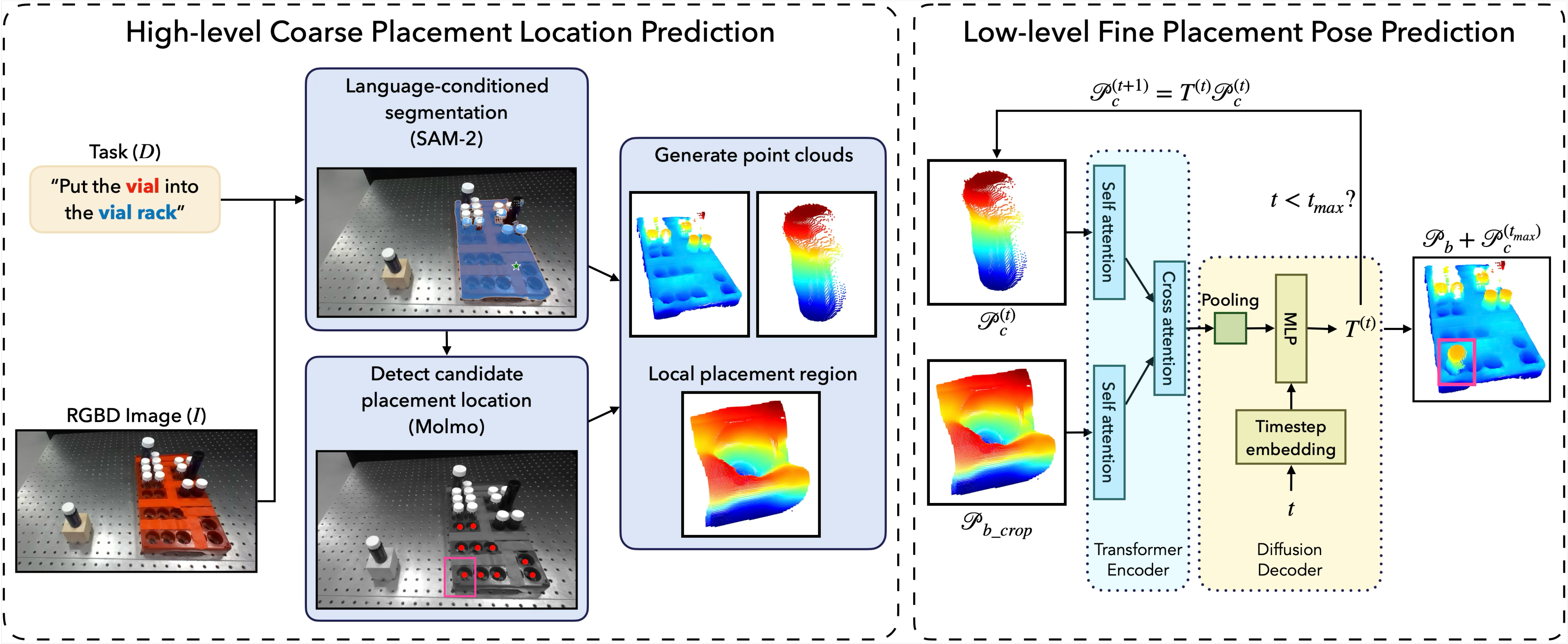}
    \caption{ \textbf{Overview of the \ourmethod placement pose prediction approach.} (1) High-level Coarse Placement Location Prediction: Given an input language description and an RGBD image, we leverage a VLM and a segmentation model to extract objects of interest. We then prompt the VLM to propose possible placement locations and crop the region of interest centered on the proposed points. The resulting point clouds are fed into the low-level model. (2) Low-level Fine Placement Pose Prediction: We use a transformer architecture, with a diffusion decoder to predict relative transformation of the point clouds to realize placement.
    %self-attention for point cloud features, cross-attention for inter-object relations, and a diffusion decoder to predict relative transformations. %(1) High-level Placement Location Prediction: Given an input language description and an RGBD image, we leverage a VLM and a segmentation model to extract objects of interest. Next, we prompt the VLM  again to propose possible placement locations. Using camera parameters, we reproject the depth map into 3D and crop the region of interest centered on the proposed placement location. Full point clouds of the objects to be placed, along with the cropped regions of the placement locations, are then fed into our low-level pose prediction model to output precise relative transformations for object placement. (2) Low-level Placement Pose Prediction: We use self-attention to extract point cloud features and cross-attention to capture cross-object features. For the diffusion decoder, the diffusion timestep is encoded within the MLP layers before the model outputs the relative transformation.
    }
    \vspace{-5pt}
\label{fig:block_scheme}
\end{figure}

\subsection{VLM-guided coarse placement location prediction}
\label{subsec:detection_pipeline}
When multiple potential placement locations exist within a task, existing models often struggle to capture all possibilities. To address this, we propose leveraging spatial VLMs, which have demonstrated strong capabilities in localizing points and regions within images based on language descriptions, to directly identify placement locations. Specifically, given a language description of the placement task \( D \) and a RGBD image \( I\), we extract the point cloud of the target object \( \mathcal{P}_{\text{c}}\), and for each identified placement point in the image, we extract the local region of the base object \( \mathcal{P}_{\text{b\_crop}}\) guided by the 3D bounding box of the target object, where \( \mathcal{P}_{\text{c}}, \mathcal{P}_{\text{b\_crop}} \in \mathbb{R}^{N \times 3} \). These point clouds are then used as input to the pose prediction model. 

%a local region of interest of the base object , where \( \mathcal{P}_{\text{c}}, \mathcal{P}_{\text{b\_crop}} \in \mathbb{R}^{N \times 3} \). 

This approach enables the low-level pose prediction model to focus on learning different placement configurations of two objects and predicting placement poses. Also, explicitly identifying placement modes, rather than relying on models to explore placements across the entire object, is more reliable and practical when handling diverse objects with multiple possible placement poses. Since our high-level module is built on a general-purpose VLM, the system can also handle diverse placements and perform complex language conditioning. We utilize Molmo \cite{deitke2024molmopixmoopenweights} to detect all potential placement locations as keypoints in image space, such as specifying all positions where a vial can be inserted into a vial plate.
% Notably, other spatial VLMs such as Gemini \cite{geminiroboticsteam2025geminiroboticsbringingai} could also be used, but are not openly available at the time of writing. 
However, our approach is not specific to one VLM, and other models with capabilities to give point locations in an image could be used \cite{geminiroboticsteam2025geminiroboticsbringingai}.
(See \appendixref{appendix:prompt} for language prompts.)

%
%First, we extract segmentation masks for the target object \( C \) and the base object \( B \). To achieve this, we first query the VLM to get point locations in the image for each object, then pass those points to the segmentation model to get segmentation masks. This allows us to have complex language conditioning in our object selection (e.g., "blue vial", "vial rack in front of the scale", etc.). Next, we query the VLM again on the base object \( B \) to find all the discrete modes for the placement. For each identified point in the image, we extract the local region in the point cloud and use that as input to our pose prediction model.
%
% (object being placed, object placement is placed onto, and specific location on that object).

% \mb{Simplification of the problem also allows us not to need diffusion -- faster training, faster inference, what else? We can achieve good results with pure transformer models, which wasn't the case previously. We can also then get additional benefits from using such a model -- energy model -- distribution of viable rotations inside one placement mode location}
% {\color{red}
% \begin{enumerate}
%     \item Language conditioning in the modes themselves (row 3, column 4)
%     \item Smaller input: better precision
%     \item Local input: generalization (which allows us to train on synthetic data)
%     \item More reliably capturing all modes
% \end{enumerate}
% }

\subsection{Fine-grained placement pose prediction}
\label{subsec:placement_model}
Given point clouds \( \mathcal{P}_{\text{c}} \) and \( \mathcal{P}_{\text{b\_crop}} \) from the high-level module, the low-level pose prediction only focuses on learning different local placement arrangements, without the need to capture the distribution of different discrete placement locations. Our intuition is that, with the aid of our large synthetic dataset, the model should effectively capture key representations of diverse placement configurations based on object geometry, which enables it to generalize to unseen objects and remain robust to noisy data.  Having only a local region as input, the pose prediction model should be able to achieve better precision, which is crucial in many relevant placement tasks.

%From the high-level module, given the two point clouds  \( \mathcal{P}_{\text{c}} \) and \( \mathcal{P}_{\text{b\_crop}} \), the model then outputs a relative transformation, comprising the translation vector and rotation matrix needed to move the placing object from its initial pose to its final placement pose. By randomly transforming initial point clouds \( \mathcal{P}_{\text{c\_transformered}}\) and use it as the model input, the model predicts a set of relative transformations \( \{T_n\}_{n=1}^{N} \) that lead to stable object placements.  

%While VLM-based local placement region extraction makes it so that our placement pose prediction model does not need to encode distribution across several, potentially distant placement locations, we still want to utilize a model that will allow us to encode a distribution of potential placement angles at the location as well as fine-grained distribution over the location itself.

 We predict the relative transformation using a diffusion model, which takes as input the two point clouds, \( \mathcal{P}_{\text{c}} \) and \( \mathcal{P}_{\text{b\_crop}} \), and through iterative denoising produces the transformation to be applied to \( \mathcal{P}_{\text{c}} \) in order for it to be placed correctly in the \( \mathcal{P}_{\text{b\_crop}} \) region. We use a transformer architecture for the encoder \cite{10.5555/3295222.3295349, chen2022neural}, where the features of the two point clouds are first extracted through self-attention layers, before being combined through cross-attention, and pooled into a latent embedding. We use a diffusion architecture for the decoder, which, conditioned on this latent embedding and the diffusion timestep, produces the delta to be applied to the object pose through transformation \( {T}_{n}^{(t)} \). For each successive diffusion step, we apply this transformation to the original point cloud \( \mathcal{P}_{\text{c}} \), in effect shifting the object being placed closer to the goal. The resulting transformation to go from the original point cloud, where the object currently is, to the final placement pose is the product of the output from each diffusion step \( T_n = \prod_{t=1}^{t_{\max}} T_n^{(t)} \). At each diffusion step during training, given the ground truth \({T}_{n, GT}^{(t)} \) and the predicted \({T}_{n}^{(t)} \), we use the L1 distance for the translation loss, the geodesic distance for the rotation loss, and apply the Chamfer loss between point clouds transformed by predicted and ground truth poses. The total loss is the sum of these individual losses. (More details in \appendixref{appendix:model}.)

\textbf{Robot pick and place execution}
After determining the placement poses, we implement a pick-and-place pipeline to manipulate the object and position it accurately at the target pose. Specifically, we utilize AnyGrasp \cite{fang2023anygrasp} to find viable grasps for the target object \(C\) and employ cuRobo \cite{curobo_report23} as the motion planner to perform collision-free placement. We perform rejection sampling on $(T_{\text{place}}, T_{\text{pick}})$ pairs to identify valid grasps for the specific placement pose predicted by our model that can be executed by the robot. Details of implementation can be found in \appendixref{appendix:eval}.

%Grasp detection begins by extracting the target object point clouds \( \mathcal{P}_{\text{c}} \) from an RGBD image. AnyGrasp \cite{fang2023anygrasp} then processes the resulting point clouds to identify the optimal grasp candidates sorted by confidence. To pick up the object, the gripper is first moved to a pre-grasp pose, positioned 10 centimeters away from the target along the gripper's z-axis. The gripper then approaches the target in a straight line while maintaining its orientation. Similarly, during placement, the robot first moves to a pre-place pose, followed by a final approach without altering the gripper orientation. The distance from the pre-place pose to the final placement pose is adjusted according to the object's size to avoid collisions during the transition to the pre-place position. With the waypoints and end-effector orientation constraints defined, we use cuRobo to generate the complete motion plan for robot pick and place.  

\section{Synthetic Dataset Generation}
\label{subsec:synthetic_dataset_generation}
\begin{figure*}[!ht]
    \vspace{-2mm}
    \centering
    \includegraphics[width=0.82\textwidth]{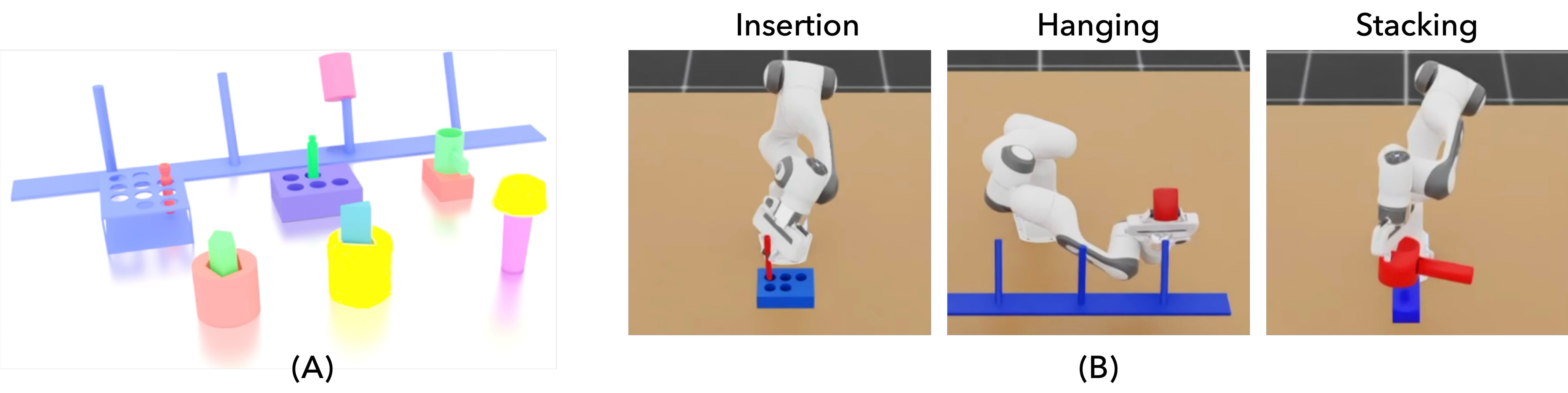}
    \vspace{-2mm}
    \caption{\textbf{Dataset generation and robot performing various placement tasks in simulation.} %In the simulation, given a set of predicted placement poses and grasping poses, we leverage cuRobo for motion planning and execute all trajectories simultaneously in IsaacLab.
    }

    \label{fig:example_placements}
    \vspace{-5mm}
\end{figure*}

Our aim in building the synthetic dataset is to capture a broad range of local placement arrangements.
Existing models use a few very specific tasks (e.g., inserting a book into a bookshelf) to simply evaluate how well their model works given placement data for such task for training \cite{yuan2023m2t2, 10160736, you2021omnihang}. Our goal, on the other hand, is not only to use the dataset to evaluate the proposed model, but to build towards representing a broad range of types of placements (stacking, hanging, inserting) as shown in \autoref{fig:example_placements}.
The local nature of our pose-prediction model makes this task much easier and enables us to build a dataset that can generalize to a broad range of real-world placement tasks.

% \begin{wrapfigure}{r}{0.4\textwidth}  % r = right, l = left
%     \vspace{-10pt}  % Optional tweak to adjust vertical alignment
%     \centering
%     \includegraphics[width=\linewidth, trim={0 10mm 0 0}, clip]{fig/synthetic_dataset_isaac_sim_high_exposure.png}
%     \caption{\textbf{Generated objects in the synthetic dataset.} We use Blender to procedurally generate objects for insertion, stacking, and hanging.}
%     \label{fig:dataset}
%     \vspace{-10pt}  % Optional to pull text up after caption
% \end{wrapfigure}

The data generation pipeline consists of two main components: object generation and placement pose generation. Specifically, we use Blender to procedurally generate 3D objects with random object parameters to further enhance diversity. To identify stable placements for objects, we use NVIDIA IsaacSim to determine object placement poses for three configurations: stacking, inserting, and hanging. This dataset covers a wide range of placement scenarios encountered in real life. In total, \SI{1489}{} objects across \SI{13}{} categories were created, and \SI{5370}{} placement poses were generated. (For more details, see \appendixref{appendix:data_gen}.)

\section{Experimental Evaluation}

% \mb{"We structure our evaluations in the following way [questions we aim to answer]:"}

% {
% \color{blue}
% \begin{itemize}
%     \item Two main evaluations:
%     \begin{enumerate}
%         \item On our diverse data
%         \item On RPDiff data -- show that our local cropping allows us: better precision, better coverage, (better success rate?), (more advanced language conditioning? -- top branch of the rack, top left placement position in shelf). Also smaller region -- smaller point cloud, faster training. Potentially also show that we don't need diffusion then. We also don't need those adaptive cropping fixed parameters that can depend on task.
%         %\item Maybe one specific dataset for precision? (Peg-in-hole)
%     \end{enumerate}
%     %\item Show an image for complex scene where we mark all the placements we execute in one run to showcase how what we do is so much more than other works. (Maybe initial and final image of the scene?). Explicitly say that people should see the video.
% \end{itemize}
% }

% \mb{say that we use the same grasp/motion plan/rejection sample framework for all methods}

We conduct evaluations against three baseline models on different placement tasks in both simulation and real-world settings. We aim to answer the following questions: 
\vspace{-1mm}
\begin{enumerate}
\item How well does \ourmethod perform in terms of object placement success rate, coverage, and precision compared to baseline models?
\item How does each component of \ourmethod described in \autoref{sec:method} affect the overall performance?
\item How well can \ourmethod generalize to novel objects and unseen configurations in the real world in a zero-shot setting, despite being trained only on synthetic data?
\end{enumerate}
\vspace{-1mm}

\begin{table}[htbp]
    \small
    \sisetup{detect-weight,     % <--
         mode=text,         % <--
         table-format=-0.4, % <--       
         add-integer-zero=false,
         table-space-text-post={*} % <--
         }
    \centering
    \caption{ \textbf{Success rate (\%) on synthetic dataset.} Overall, \ourmethod achieves consistently high success rates across four pick-and-place tasks on the synthetic dataset.
    }
    \label{tab:sim_sr}
    \resizebox{\linewidth}{!}{
        \begin{tabular}{l | l | S[table-format=3.2] S[table-format=3.2] | S[table-format=3.2] S[table-format=3.2]}
        \toprule
            &  \textbf{Methods}
            & \textbf{Object Stacking}
            & \textbf{Peg Insertion}
            & \textbf{Cup Hang}
            & \textbf{Vial Insertion} 
            \\
            &
            & \textbf{(single-mode)}
            & \textbf{(single-mode)}
            & \textbf{(multi-mode)}
            & \textbf{(multi-mode)} 
            \\
        \midrule
        \textbf{Single task} & 
        NSM~\cite{chen2022neural}&                            76.57 & 7.63 & 35.54 & 18.70 \\
        \textbf{training} &  RPDiff~\cite{simeonov2023rpdiff} &                           \B 80.34 & 22.94 & 92.02 & 16.51 \\
        & AnyPlace-EBM (ours) &             \B  80.04 & 8.44 & 91.57 &65.64 \\
        &  \textbf{\ourmethod} (ours) &            \B  80.16 &  \B 30.95 & \B 94.80 &  \B 92.74 \\
        \midrule
        \textbf{Multi-task} & NSM~\cite{chen2022neural}  &                  77.55 & 7.69 & 35.22 & 9.87 \\
        \textbf{training} &  RPDiff~\cite{simeonov2023rpdiff}  &                 \B 80.21 & 22.33 &   \B 94.05 & 24.26 \\
        & AnyPlace-EBM (ours)  &     78.95 & 10.75 & 90.87 & 57.24 \\
        &  \textbf{\ourmethod} (ours)  &     78.28 &  \B 24.99 &   \B 94.12 &  \B 75.25 \\
        \bottomrule
        \end{tabular}
        }
    \vspace{-2mm}
\end{table}

\noindent \textbf{Evaluation metrics.}
We use three metrics to evaluate model performance: success rate, coverage, and precision. Specifically, we define a placement as successful if the robot places the object at the correct location, and it remains stable after release. The success rate is then calculated as the number of successful placements divided by the total number of trials. To better understand the diversity of multimodal outputs in placement prediction, we evaluate coverage, defined as the number of distinct predicted placement locations relative to the total number of possible locations. Finally, for fine placement tasks, to evaluate the precision, we measure the error between the ground truth pose and the predicted pose in terms of both distance and angle.

% \subsection{Baselines}
\noindent \textbf{Baselines.}
We compare our approach with three baselines: NSM \cite{chen2022neural}, RPDiff \cite{simeonov2023rpdiff}, and an energy-based model (EBM), a variant of our model that is integrated with our high-level placement location prediction module. Specifically, NSM shares the same self-attention and cross-attention point cloud encoder, paired with a regression decoder. For RPDiff, since the low-level pose-prediction module in our diffusion-based approach shares the same structure, this allows us to directly examine the effects of the high-level module we propose.
%We do not utilize a learned classifier on top of the pose prediction model that is present in RPDiff. Such a model can be applied on top of any of the methods we evaluate here, including the ones we propose. Not having it also allows us to evaluate the number of samples needed to achieve particular coverage of possible placement locations in the scene.
To evaluate the effectiveness of the diffusion decoder in generating multimodal outputs, we build AnyPlace-EBM inspired by Implicit-PDF \cite{implicitpdf2021}, where the model uses the same encoders as ours and is trained to assign low energy values to stable placement poses (Details in \hyperref[appendix:energy]{Appendix \ref{appendix:energy}}). %More deception of baseline models can be found in the Appendix. %Inspired by Implicit-PDF \cite{implicitpdf2021}, this model uses the same encoder as ours, but for the decoder, instead of explicitly predicting the placement pose, it includes two separate branches: one for placement location prediction and another for predicting the placement rotation energy. During training, we encourage all placement rotations in SO(3) that result in stable placements to have low energy. During inference, we randomly sample thousands of rotations and select the one with the lowest energy as the final placement rotation. 
Each model is trained independently on placement-type-specific subsets and on the full dataset (the multitask variant).

\textbf{Placement Success: Single \& Multi-Modal.} We evaluate all methods in simulation using our pick-and-place execution pipeline within IsaacLab \cite{mittal2023orbit}. Experiments cover four tasks: \texttt{object stacking} and \texttt{peg insertion} (single-mode), and \texttt{cup hang} and \texttt{vial insertion} (multi-modal, with multiple valid placements). A summary of results is shown in \autoref{tab:sim_sr}.

In single-mode tasks, for the simple stacking task, where precision in placement poses is not required, all models achieve a similar success rate. For the peg-in-hole task, which requires high precision in placement pose prediction, \ourmethod surpasses the baseline models by a large margin. Multimodal tasks are where we expect to get the full benefit from our approach. In the hanging task, the tolerance for placing a cup on the rack is relatively high, and \ourmethod outperforms baseline models slightly. In the more challenging vial insertion task, \ourmethod achieves the highest success rate of 92.74\%, while the success rates of all baseline models drop significantly. This demonstrates that relying on high-level VLM to propose possible placement locations and focusing solely on the local region for placement prediction simplifies the task for low-level pose prediction models and enables them to better capture fine-grained point cloud features for high-precision placements. Additionally, despite using the same high-level placement location prediction, the energy-based model suffers a worse performance compared to the diffusion-based \ourmethod. This highlights that the iterative denoising procedure in diffusion is more effective for high-precision placement prediction.

\renewcommand\thesubfigure{\Alph{subfigure}}  % Capital letters for subfigure labels

\begin{figure}[t]
    \centering
    \begin{subfigure}[t]{0.45\textwidth}
        \centering
        \includegraphics[trim=10 10 20 10, width=\linewidth]{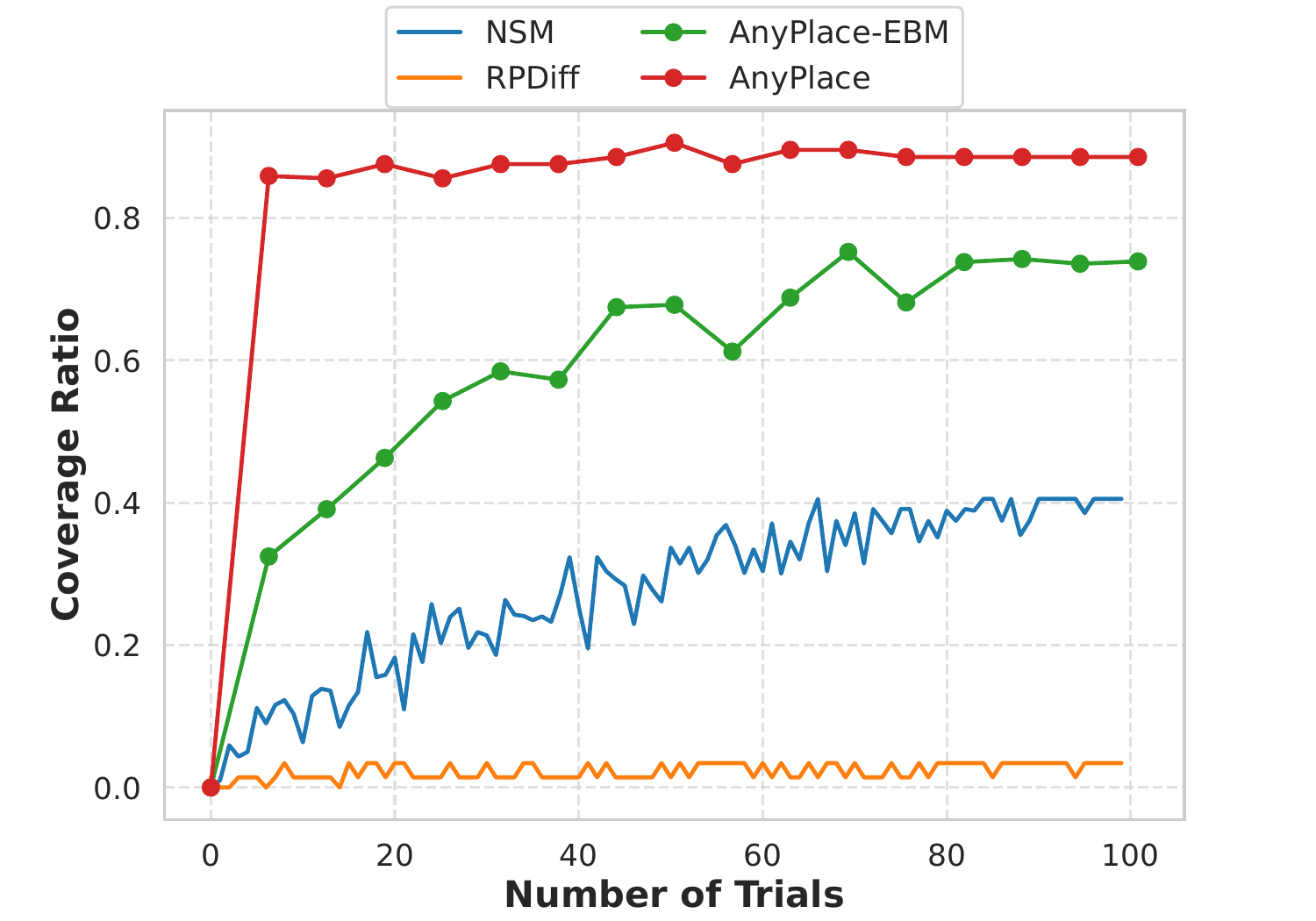}
        \caption{Vial Insertion Task}
        \vspace{-2mm}

        \label{fig:vialinsertion_coverage}
    \end{subfigure}
    \hfill
    \begin{subfigure}[t]{0.45\textwidth}
        \centering
        \includegraphics[trim=10 10 20 10, width=\linewidth]{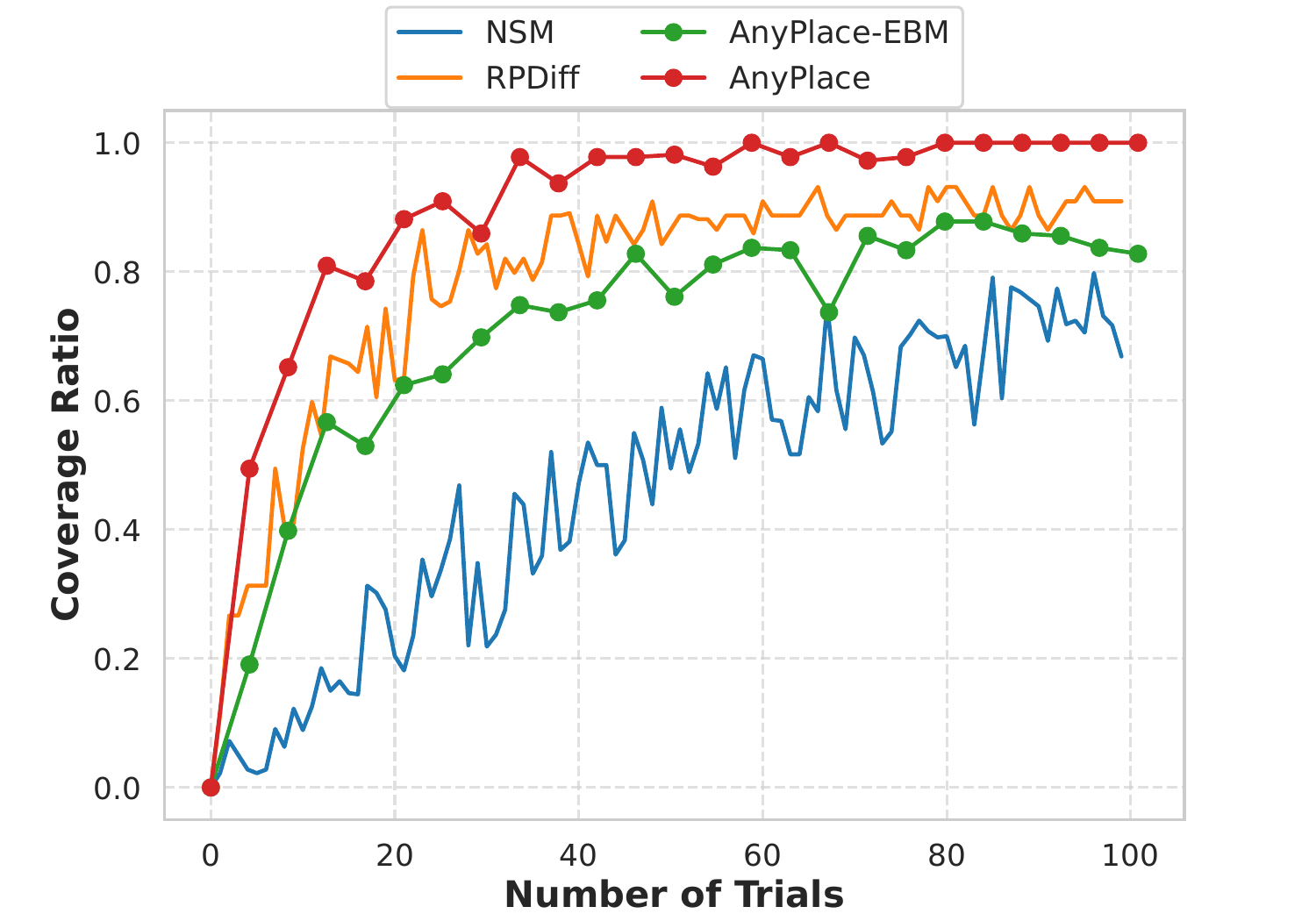}
        \caption{Hanging Task}
        \vspace{-2mm}
        \label{fig:hanging_coverage}
    \end{subfigure}
    \caption{\textbf{Coverage comparison across different models in vial insertion and hanging tasks.} In both cases, \ourmethod achieves near-perfect coverage with just a few samples, while baselines fail to match this even with 100 samples.}
    \label{fig:coverage_comparison}
    \vspace{-2mm}
\end{figure}

% \begin{figure}[t]
%     \begin{center}
%     \begin{minipage}[t]{0.69\textwidth}
%         \begin{subfigure}[t]{0.5\textwidth}
%             \centering
%             \includegraphics[trim=0 0 60 50, width=\linewidth]{fig/insertion_new.pdf}
%         \end{subfigure}
%         \hfill
%         \begin{subfigure}[t]{0.5\textwidth}
%             \centering
%             \includegraphics[trim=0 0 60 50, width=\linewidth]{fig/hanging_new.pdf}
%         \end{subfigure}
%         \caption{
%             \textbf{Coverage comparison across different models.} \textbf{Left:} Vial insertion task. \textbf{Right:} Hanging task.
%         }
%         \label{fig:coverage_comparison}
        
%     \end{minipage}
%     \hfill
%     \begin{minipage}[t]{0.28\textwidth}
%         \begin{table}[H]  % float placement 'H' requires \usepackage{float}
%             \centering
%             \vspace{-30mm}
%             \scriptsize
%             \label{tab:hanging_success_rate}
%             \begin{tabular}{@{}l|c}
%                 \toprule
%                 Methods & Success Rate \\
%                 \midrule
%                 NSM               & 37.26\% \\
%                 RPDiff            & 81.17\% \\
%                 AnyPlace-EBM      & 69.43\% \\
%                 \textbf{AnyPlace} & \textbf{84.96\%} \\
%                 \bottomrule
%             \end{tabular}
%             \vspace{10mm}
%              \caption{\textbf{Hanging success rate of multitask models.}}
%         \end{table}
%     \end{minipage}
%     \end{center}
% \end{figure}

\textbf{Placement Coverage: Multi-Modal Insertion \& Hanging.} We now investigate how coverage changes as the number of samples taken from each model increases.
Given the design of NSM and RPDiff, we can only take independent samples from the model until all possible placement modes are covered.
In contrast, for \ourmethod and \energyvlm, the high-level VLM module predicts possible placement locations, allowing us to perform sampling for each mode separately.
We perform an equal number of samples for each mode and compare performance based on the same total number of samples across all models.

In \autoref{fig:vialinsertion_coverage}, we show results for the vial insertion task. \ourmethod rapidly approaches peak performance with a single sample per mode due to its high placement success rate and the VLM's strong reasoning in identifying diverse placement locations. \energyvlm models show a similar trend but plateau at 73\% due to lower success rates. In contrast, RPDiff fails to capture multimodality, with coverage below 10\%, and is outperformed even by the NSM regression model. For hanging, we assess generalization across racks with varied sizes, geometries, and stick spacing. The placement coverage is reported in \autoref{fig:hanging_coverage}. As with vial insertion, \ourmethod consistently achieves 100\% coverage with fewer samples. While RPDiff improves here, its coverage saturates at 90\%, revealing limited generalization. In contrast, \ourmethod’s coarse predictions guide the fine model to focus on local placement, enabling robust generalization across object variations.

\begin{figure*}[t]
    \centering
    \vspace{-3mm}
    \begin{minipage}[t]{0.5\textwidth}
        \centering
        \includegraphics[width=0.95\linewidth]{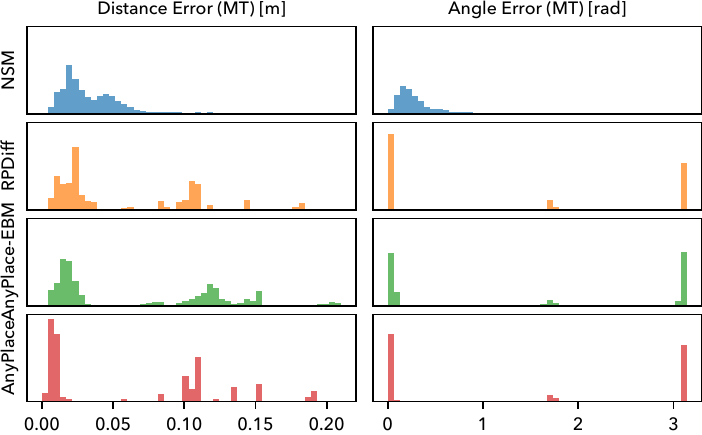}
        \caption{
            \textbf{Errors on insertion tasks.} Based on histograms of the translation and rotation error in pose prediction, it is clear that AnyPlace achieves the best precision. (MT): models trained on multi-task data.
        }
        \label{fig:precision-combined}
    \end{minipage}%
    \hfill
    \begin{minipage}[t]{0.47\textwidth}
        \centering
        \vspace{-33mm}
        \scriptsize
        \begin{tabular}{@{}l|c|c|c}
            \toprule
            Methods & Insert vial & Hang ring & Stack battery\\
             & (10 modes)  & (5 modes) & (3 modes)\\
            \midrule
            NSM               & 0\% & 0\% & 0\% \\
            \rowcolor{gray!15} RPDiff            & 0\% & 60\% & 0\% \\
            AnyPlace-EBM      & 50\% & 0\% & 0\% \\
            \rowcolor{gray!15} \textbf{AnyPlace} & \B 80\% & \B 80\% & \B67\% \\
            \bottomrule
        \end{tabular}
        \vspace{5mm}
        \captionof{table}{
            \textbf{Coverage rate of the real robot executing three placement tasks.} 10 trials were conducted for each task. Our model significantly outperforms baseline models in real experiments, demonstrating its ability to generalize to unseen objects and effectively handle noisy data.
        }
        \label{tab:real_sr_data}
    \end{minipage}
    %\vspace{-4mm}
\end{figure*}

\textbf{Placement Precision: Fine-grained Insertion.} Our final simulation evaluation focuses on assessing precision, a key factor in many placement tasks. We test whether restricting input regions improves placement pose precision by directly analyzing model predictions. \autoref{fig:precision-combined} shows distance and rotation error distributions for insertion tasks. We base the error on predicted poses to the closest viable placement pose. \ourmethod yields smaller and more consistent errors than baselines. We also evaluate rotation predictions, excluding yaw due to object symmetry. As the initial pose is randomized and models cannot determine this orientation, all methods produce uniformly random yaw angle values. Most models predict flipped and correct orientations equally, except NSM, which performs worse in both position and orientation. Both \ourmethod and RPDiff perform well on orientation, while AnyPlace-EBM achieves slightly lower accuracy than RPDiff but surpasses NSM, highlighting its potential as a non-diffusion-based alternative.

\begin{figure*}[t]
    \vspace{-2mm}
    \centering
    \includegraphics[width=0.67\textwidth]{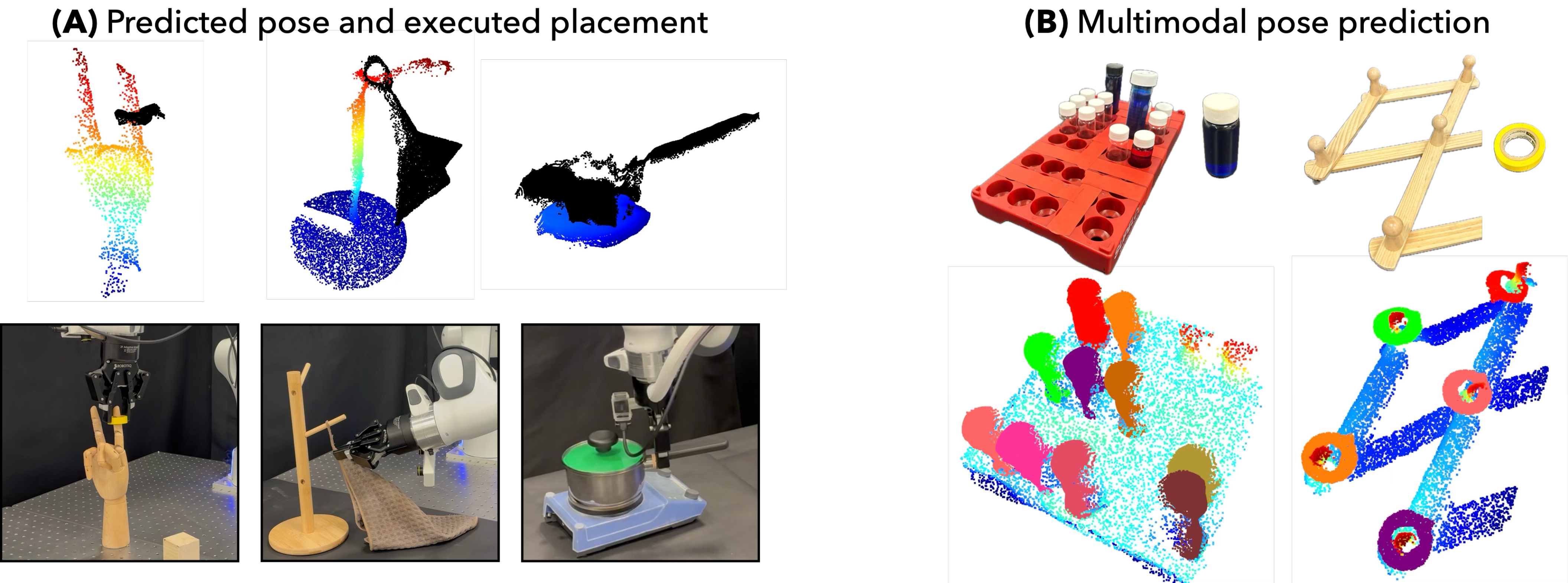}
    \vspace{-2mm}
     \caption{\textbf{Demonstration of robot executing diverse real-world tasks using AnyPlace predictions.} Trained only on synthetic data, the model generalizes to unseen objects and effectively handles noisy point clouds.
    }
\label{fig:real_pcd}
    \vspace{-2mm}
\end{figure*}

\subsection{\textbf{Real World Evaluation}}

We evaluate our approach on \numtasks real-world placement configurations with different objects. For each scene, a single RGBD image is captured using a ZED Mini camera mounted on a Franka Emika arm. We use the same high-level pipeline and models trained solely on synthetic data to predict placement poses in a zero-shot setting. We use a simplified pipeline for executing the placement, using a specific grasp and performing placement inverse kinematics instead of a full motion planner. In addition, we do not utilize rejection sampling, but directly execute each trajectory on the real robot. Our goal is to test whether predicting local placements enables generalization to unseen objects. Additional visualization and discussion can be found in \appendixref{appendix:eval}.

\textbf{Precise Multi-modal Placement}. 
We perform systematic evaluations of coverage and success rate over 10 trials on three tasks, as shown in \autoref{fig:teaser}, each involving multiple possible placement locations, as shown in \hyperref[tab:real_sr_data]{Table \ref{tab:real_sr_data}}. Overall, NSM and RPDiff fail to adapt to the differences in real-world tasks compared to the training conditions. \ourmethod, on the other hand, successfully completes placements, achieving the best performance across all tasks --- 80\% on the fine-grained vial insertion task. This demonstrates not only the effectiveness of the VLM in identifying placement modes, but also the generalization and precision of our low-level pose prediction model.

\textbf{Generalizable \& Language-conditioned Placement}. 
To evaluate generalization, we extensively test \ourmethod on a variety of real-world placement tasks. The robot successfully performs diverse placements with unseen rigid and deformable objects, such as stacking a funnel on a holder, hanging tools and a towel on racks (\hyperref[fig:teaser]{Figure \ref{fig:teaser}A-E}). It handles fine-grained (\hyperref[fig:teaser]{Figure \ref{fig:teaser}F-H}: peg-in-hole, ring stacking) and long-horizon tasks (\hyperref[fig:teaser]{Figure \ref{fig:teaser}N-O}: lid-on-pot, followed by pot-on-stove). It can also perform placement at different locations based on language description (\hyperref[fig:teaser]{Figure \ref{fig:teaser}K-M}: bottle on different shelves). We also visualize raw pose predictions made by the model for a selection of these experiments, by moving the object point cloud to the predicted pose (\hyperref[fig:real_pcd]{Figure \ref{fig:real_pcd}A}), as well as showing multimodal predictions at the same time (\hyperref[fig:real_pcd]{Figure \ref{fig:real_pcd}B}).
%
%In \autoref{fig:real_pcd}, we show object point clouds at predicted poses and execution of those same predictions on the real robot (), as well as point clouds for all predictions from the model in multimodal cases. also visualize raw object point clouds at predicted poses and real robot execution. 
These experiments demonstrate that combining VLM-based location prediction with local pose refinement enables accurate and language-guided placements. This integration, along with synthetic training, allows generalization to unseen objects and complex configurations.

\section{Conclusion}
In this work, we presented a general pipeline for performing a wide range of object placement tasks using a robotic arm. We proposed a two-part framework consisting of a high-level module that determines coarse placement locations and a low-level module that predicts fine placement poses. The core idea of our approach is to leverage a VLM to propose placement locations, allowing the low-level pose prediction model to focus only on the local region of interest in the object’s point cloud. This effectively reduces complexity and enhances generalization. To train our model, we created a synthetic dataset containing thousands of randomly generated objects and placement poses. We demonstrated the effectiveness of the entire pipeline in both simulation and real-world experiments. In simulation, we showed that \ourmethod outperforms baseline methods in terms of success rate, coverage, and precision. We then validated its robustness and generalization in real-world settings, where, given a single RGB-D image, \ourmethod predicts diverse placement configurations in a zero-shot manner and successfully generalizes to unseen objects, despite being trained purely with synthetic data.

% \mb{
% VLM models with strong pixel level grounding are just starting to appear, with the first one being able to enable a pipeline like we propose being Molmo. Even with it our high level is pretty powerful, it will only get better.
% }

\clearpage

\section{Limitations}

% \mb{Do a pass.}

% In this work, we focus solely on predicting the relative transformation of objects to be placed. To execute the pick-and-place task, pre-place poses need to be manually defined, which adds an additional step. Moreover, rejection sampling in real-world systems can be time-consuming, preventing the system from operating in real time. Exploring an end-to-end pick-and-place policy could be a promising direction for future work.

While our main focus in this work is on predicting placement poses, there are still challenges to be tackled in order to be able to execute the full pick-and-place task with the same generality. Not every stable grasp of an object can be used to realize the placement of the same object at a specific pose, and performing rejection sampling in real-world scenarios can be difficult and time-consuming. We do believe our synthetic dataset and evaluation pipeline provide a great foundation for making progress in this direction, by using them to generate data for training an end-to-end pick-and-place model applicable to a wide range of real-world placement tasks.

A common limitation of object rearrangement approaches, similar to AnyPlace, is their lack of consideration for object contact interactions during placement. By only controlling the robot end-effector placement pose based on our model prediction, the robot approaches objects too aggressively, leading to hard impacts or failed placements due to a lack of smooth and soft contact. Nevertheless, we demonstrate that AnyPlace is capable of handling a diverse set of placement tasks in the real world. A promising direction for future work is to incorporate compliance control or force feedback to adapt the robot's motion during contact and achieve more reliable placements.

While our approach enables greater precision in the placement pose prediction, it is still limited by the precision of the point clouds it receives as input.
Completing placement tasks requiring significant precision with imperfect point cloud data can be very difficult. Specifically, we visualize the object point clouds at the predicted placement poses for failure cases (see Appendix \autoref{fig:real_predict_fail}).
For ring hanging, the point cloud of the rack is too sparse, causing the predicted position to be off, although the orientation remains mostly correct. For peg inserting, the captured point clouds of the holes are noisy and have vague boundaries. While our model predicts the correct orientation, it slightly misses the hole, leading to insertion failure. For battery stacking, the point cloud of the battery is largely incomplete, causing the predicted placement to be tilted rather than vertical. The battery falls over after the robot releases it.
For future work, recent methods for estimating depth from RGB images have some potential to aid in this, as does continued progress in sensor quality.
Another very promising approach for tackling this problem is having a policy for performing the final section of the placement based on force/torque feedback.
Similarly, like in the previous point, we also think that our dataset and simulation pipeline can be a great starting point for tackling this issue, by providing us with data for training such reactive placement-execution policies.

The approach we are proposing also does not include language conditioning in the low-level pose prediction model. This means that we are unable to choose between different types of placements in the same location. However, with the architecture of the low-level model, it is straightforward to add an additional language-conditioning input. Our synthetic data generation pipeline lends itself well to data generation in this case as well, where we can automatically add language conditioning to different placement types and even use an LLM to add diversity to language-conditioning data.

% In addition, during real-world experiments with the peg-in-hole task, we observed that noise in the hole object point cloud caused inaccuracies in the orientation of the predicted peg pose, leading to failures. As a future direction, incorporating images as input could help address this issue by capturing better geometry information and extracting richer visual representations of objects.

% \mb{
% \begin{enumerate}
%     \item In the approach we proposed there is no language conditioning in the low-level pose prediction. This means that we are unable to choose between different types of placements in the same location. However, with the architecture of the low-level model, it is simple to add an additional language-conditioning input. As for the data for learning this additional language conditioning, our synthetic data generation pipeline lends itself great to automatically be able to generate it, and even use LLM to add diversity to language-conditioning data.
% \end{enumerate}
% }

% The acknowledgments are automatically included only in the final and preprint versions of the paper.
\acknowledgments{This research was
undertaken thanks in part to funding provided to the
University of Toronto’s Acceleration Consortium from
the Canada First Research Excellence Fund, grant number CFREF-2022-00042. We acknowledge the generous
support of Dr. Anders G. Fr{\o}seth, the Acceleration Consortium, the Vector Institute, Natural Resources Canada
and the Canada 150 Research Chairs program.}

%===============================================================================

\bibliography{main}  % .bib

\clearpage  
\appendix
\renewcommand{\thefigure}{A\arabic{figure}}
\setcounter{figure}{0}
% \section*{Appendix}
\section{Language Prompts for Molmo and Predicted Placement Visualization}
\label{appendix:prompt}
\begin{figure*}[!h]
    \centering
    \includegraphics[width=0.9\textwidth]{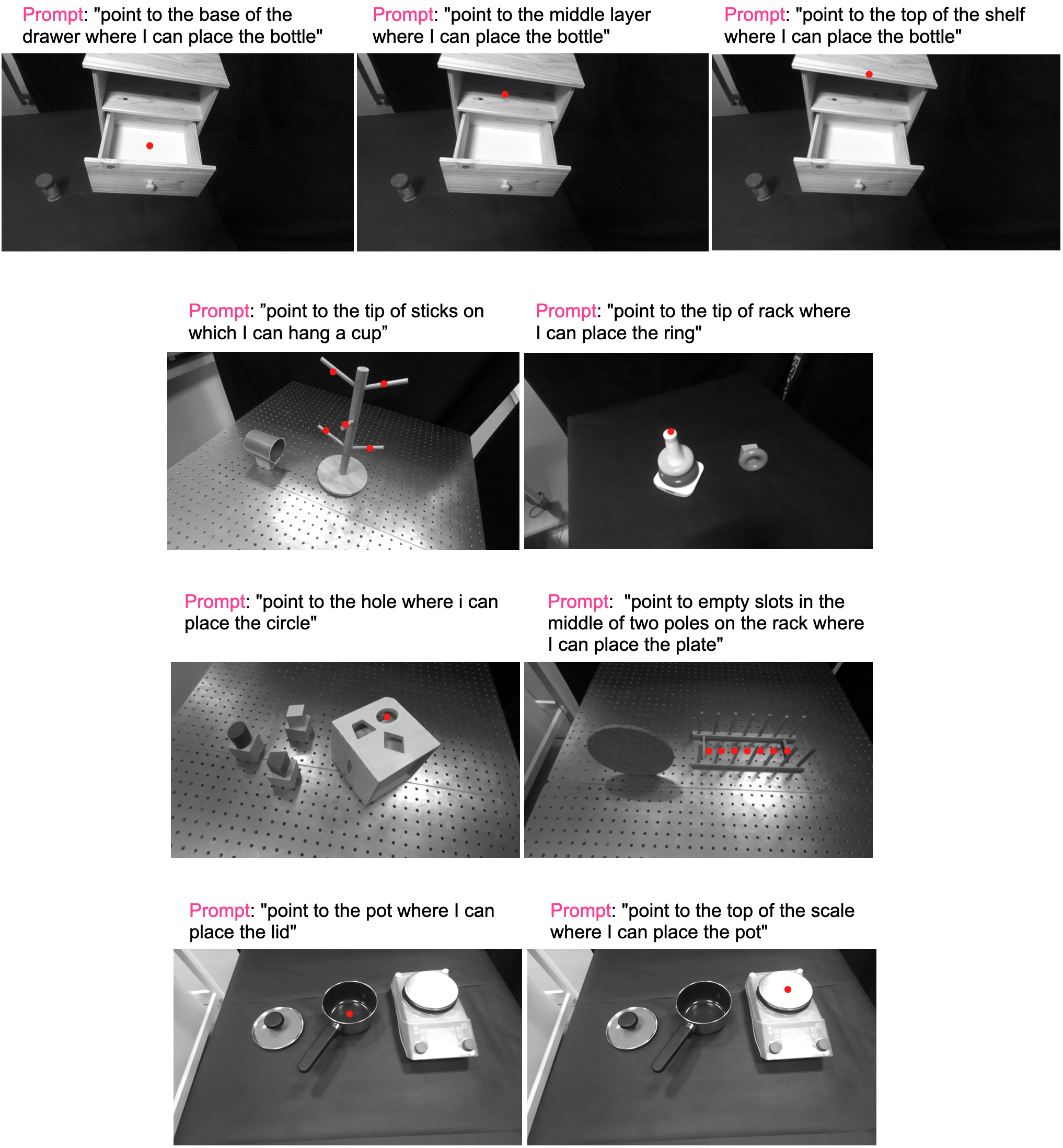}
    \caption{\textbf{Additional Language Prompts and VLM Output Visualization in the Real-World Evaluation.} }
    \label{fig:prompt}
\end{figure*}

\begin{figure*}[!h]
    \centering
\includegraphics[width=0.75\textwidth]{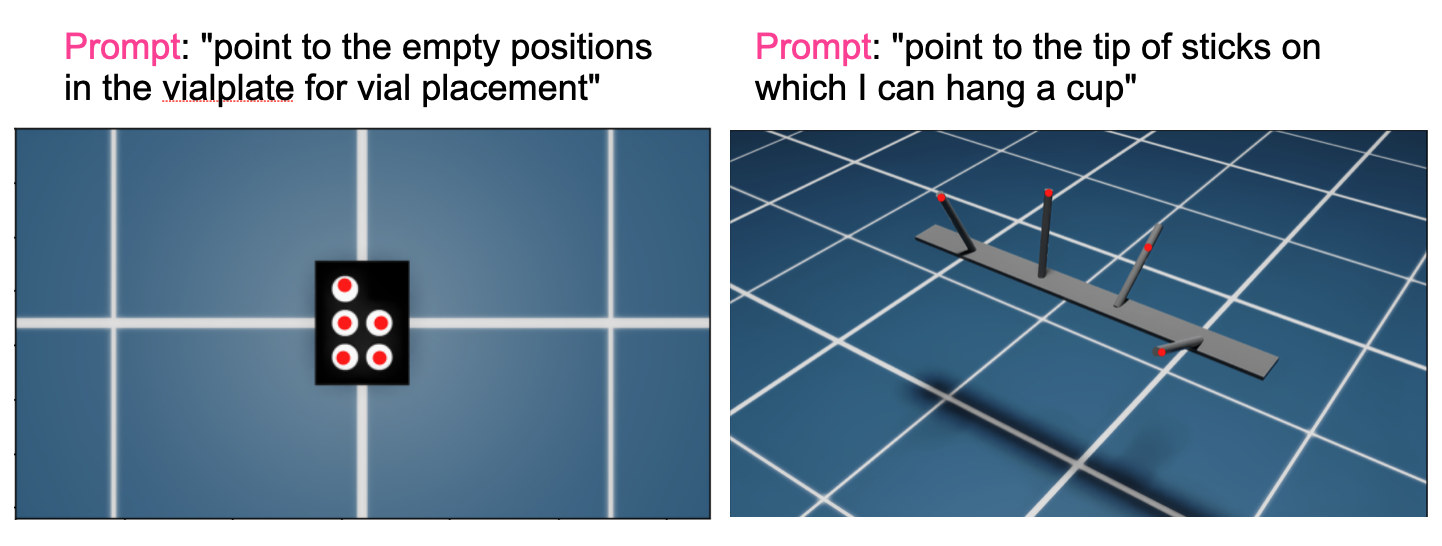}
    \caption{\textbf{Language Prompts and VLM Output Visualization in the Simulation Evaluation.} In our simulation experiments, we use the VLM to generate placement locations on RGB images from the simulator. Based on this proposed location, we crop the point clouds and input them into our placement pose prediction model.}
    \label{fig:prompt_sim}
\end{figure*}

 To extract object point clouds, we first prompt Molmo (e.g., points to the bottle) to predict a single anchor point. This anchor is then passed to SAM-2 to obtain segmentation masks of the object. The complete point clouds are subsequently extracted using the segmentation and camera information. To predict placement locations, in \autoref{fig:prompt} and \autoref{fig:prompt_sim}, we show language prompts used by the VLM in both real-world and simulation experiments. Based on our observations, the Molmo VLM accurately identifies the correct placement locations in both real-world and simulated images based on the language input. Even when the predicted location is not perfectly centered, our low-level pose prediction model can still be successful in predicting the placement pose. For instance, when predicting the placement of the bottle on the top shelf (\autoref{fig:prompt}, last image in the top row), the VLM may give a location at the very edge. However, our pose-prediction model focuses on the entire local region extracted around that point and is able to provide a pose where the entire object is on the surface, allowing the robot to execute the task successfully.

 We do notice that minor prompt engineering is occasionally needed to generate accurate predictions using Molmo, such as “point to the empty slot between the two poles”. We believe this limitation can be addressed as VLM models continue to improve.

\section{Additional Details on the Low-level Diffusion Model}
\label{appendix:model}

\subsection{Model Architecture}
For our diffusion-based placement prediction model, we leverage a transformer architecture. Details of the model architecture are listed in \autoref{tab:model_arch}. Starting with two object point clouds, the diffusion process iteratively denoises the relative transformation, gradually moving the object being placed toward its final pose. Initially, we transform the object point cloud \( \mathcal{P}_{\text{c}} \) with a random transformation \( T_{\text{init}} = (\mathbf{R}, \mathbf{t}) \) to get \( \mathcal{P}_{\text{c}}^{\text{(0)}} \), where R is randomly sampled over the SO(3) space, and t is sampled within the bounding box of the cropped placement region:
\begin{equation}
\mathcal{P}_c^{(0)} = T_{\text{init}}\, \mathcal{P}_c,
\end{equation}
where
\begin{equation}
T_{\text{init}} = (\mathbf{R}, \mathbf{t}), \quad \mathbf{R} \sim \mathcal{U}(\mathrm{SO}(3)), \quad \mathbf{t} \sim \mathcal{U}(\text{bbox}(\mathcal{P}_{b\_crop})).
\end{equation}

 As shown in \autoref{fig:block_scheme}, at each denoising timestep \( t \), \( \mathcal{P}_{\text{c}}^{\text{(t)}} \) and \( \mathcal{P}_{\text{b\_crop}}\) are input into the encoder. Specifically, both point clouds are first downsampled to \SI{1024} points using Farthest Point Sampling (FPS) and normalized to the size of a unit cube. The downsampled point clouds are passed through a linear layer to extract latent features, which are subsequently concatenated with a one-hot vector used to identify the corresponding point cloud. These combined features are then processed by the Transformer encoder \cite{10.5555/3295222.3295349, chen2022neural}, where self-attention layers are applied to effectively extract features from the point clouds. We then leverage cross-attention and pooling layers to further aggregate these features, producing a unified feature representation that captures the spatial relationship between the two objects. In the decoder, we first obtain the sinusoidal positional embedding of the diffusion timestep \(t\). Finally, the joint point cloud feature representation, along with the encoded timestep, is fed into MLP layers to predict the relative transformation \( {T}_{n}^{(t)} \) consisting of a rotation \( \mathbf{R} \in \text{SO}(3) \) and a translation \( \mathbf{t} \in \mathbb{R}^3 \) for refining the object's pose. The target object point clouds are then transformed accordingly before proceeding to the next denoising step as $ \mathcal{P}_{\text{c}}^{\text{(t-1)}} =  {T}_{n}^{(t)} \mathcal{P}_{\text{c}}^{\text{(t)}}$.
 The full transformation \( T_n \) taking the object from its initial location to the placement pose is the product of all the incremental transformations predicted through the diffusion steps: \( T_n = \prod_{t=1}^{t_{\max}} T_n^{(t)} \).

We do not utilize a learned classifier on top of the pose prediction model that is present in RPDiff. Such a model can be applied on top of any of the methods we evaluate here, including the ones we propose. Not having it also allows us to evaluate the number of samples needed to achieve a particular coverage of possible placement locations in the scene.

\begin{table*}[h]
    \centering
    \normalsize
    \caption{Summary of Model Architecture}
    \label{tab:model_arch}
    \begin{tabular}{@{}lcc@{}}
        \toprule
        \textbf{Model Component} & \textbf{Details} \\ 
        \midrule
        Model Total Parameters &  4,279,688 \\ 
        Number of Heads & 1 \\ 
        Number of Self-attention Blocks & 4 \\ 
        Number of Cross-attention Blocks & 4 \\ 
        Point Cloud Feature Dimension & 258 (256 + one-hot embedding) \\ 
        Transformer Feature Dimension & 256 \\ 
        \midrule
        \multicolumn{2}{l}{\textbf{Encoder}} \\ 
        \midrule
        Self-Attention & Multi-Headed Attention (1 head) \\ 
        Feedforward Layer & Linear (258 $\to$ 256), ReLU, Linear (256 $\to$ 258) \\ 
        Normalization & LayerNorm \\ 
        \midrule
        \multicolumn{2}{l}{\textbf{Decoder}} \\ 
        \midrule
        Self-Attention & Multi-Headed Attention (1 head) \\ 
        Cross-Attention & Multi-Headed Attention (1 head) \\ 
        Feedforward Layer & Linear (258 $\to$ 256), ReLU, Linear (256 $\to$ 258) \\ 
        Normalization & LayerNorm \\ 
        \bottomrule
    \end{tabular}
\end{table*}

\subsection{Model Training} During training of our low-level diffusion pose prediction model, we perform 5 denoising steps. Instead of incrementally adding Gaussian noise to the input during the forward process, as is common practice \cite{10.5555/3495724.3496298}, we manually define the noise added at each timestep. In our case, the noise is the relative transformation that the model predicts. Specifically, the intermediate ground truth relative transformations \({T}_{n, GT}^{(t)} \) are generated by linearly interpolating the translation and using spherical linear interpolation (SLERP) to sample rotations between the object's initial and final placement poses. During inference, we generate diverse placement poses by sampling the diffusion model multiple times, each time starting with randomly transformed initial object point clouds \( \mathcal{P}_{\text{c}}^{\text{(0)}} \). We perform a larger number of denoising steps at test time, by repeating the last denoising step more times for a total of 50 denoising steps, similar to RPDiff \cite{simeonov2023rpdiff}. All single-task models are trained for three days, while multitask models are trained for five days on a single NVIDIA A100 GPU. Other parameters can be found in \autoref{tab:model_train}.

\begin{table}[h!]
    \centering
    \normalsize
    \caption{Parameters for Model Training}
    \renewcommand{\arraystretch}{1.2}
    \label{tab:model_train}
    \begin{tabular}{@{}ll@{}}
        \toprule
        \textbf{Parameter} & \textbf{Value} \\ 
        \midrule
        Diffusion steps & 5\\
        Number of training iteration & 500k\\
        Batch size & 48 \\ 
        Optimizer & AdamW \cite{loshchilov2019decoupledweightdecayregularization}\\ 
        Learning rate & \( 1 \times 10^{-4} \) \\ 
        Learning rate schedule & Linear warmup and cosine decay \\ 
        Warmup epochs & 50 \\ 
        Weight decay & 0.1 \\ 
        Optimizer momentum & \( \beta_1 = 0.9 \), \( \beta_2 = 0.95 \) \\ 
        \bottomrule
    \end{tabular}
\end{table}

\subsection{Additional Details on Data Generation}
\label{appendix:data_gen}
For object generation in Blender, we procedurally generate 3D objects, such as pegs, holes, cups, racks, beakers, vials, and vial holders. Object parameters—including height, width, length, shapes, and number of edges—are randomized to increase variability. For racks and vial plates, we also randomize the number of poles and holes. Additionally, random scaling is applied along the x, y, and z axes. This type of programmatic object generation essentially gives us placement poses ”for free”. 

This approach for object generation not only allows simple randomization of object geometries, but also supports a range of valid placements rather than a single fixed pose. For training the AnyPlace low-level pose-prediction model, only local placement data is needed, so we do not have to generate full objects with all possible placement modes, further simplifying scalability. While adding completely new object classes does require some human effort, it is limited to writing a single piece of code per class, and common geometric aspects can often be shared. Overall, building a large dataset of parametrized object classes enables programmatic creation of vast amounts of synthetic training data, supporting not only placement tasks but potentially many other manipulation problems.

To identify stable placements for objects, we use NVIDIA IsaacSim to determine object placement poses. At the start of each trial, two objects are randomly sampled and loaded into the simulation. Since all objects are procedurally generated and possible placement locations are known during generation (e.g., the center of each hole on a vial plate), the ideal object placement location for the placing object can easily be determined. This approach finds object placement locations that maximize the clearance between the objects in their final placement configurations. For object placement rotations, they are then randomly sampled along their axis of symmetry to explore various placement poses.  Four cameras are set up to capture dense object point clouds and render RGBD images. This dataset covers a wide range of placement scenarios encountered in real life. \autoref{tab:dataset} presents statistics on the number of placements generated using our synthetic dataset generation pipeline. Notably, by focusing only on the region of interest for placement pose prediction, \ourmethod models perform well across different placement tasks and generalize to unseen objects. They are trained with fewer than \SI{2000} samples per task, demonstrating the effectiveness of our design.

\begin{table}[h!]
    \normalsize
    \centering
    \caption{Dataset Size for Each Placement Task}
    \renewcommand{\arraystretch}{1.2}
    \label{tab:dataset}
    \begin{tabular}{@{}l c@{}}
        \toprule
        \textbf{Placement Task} & \textbf{Number of Placements} \\ 
        \midrule
        Hanging & 1,767\\
        Stacking & 1,696\\
        Vial Insertion & 1,107 \\ 
        Other Insertion & 800\\ 
        \bottomrule
    \end{tabular}
\end{table}

\section{Evaluation}
\label{appendix:eval}

\subsection{Implementation of the Pick and Place Pipeline}
After determining the placement poses by AnyPlace, we implement a pick-and-place pipeline to manipulate the object and position it accurately at the target pose. Specifically, grasp detection begins by extracting the target object point clouds \( \mathcal{P}_{\text{c}} \) from an RGBD image. AnyGrasp \cite{fang2023anygrasp} then processes the resulting point clouds to identify the optimal grasp candidates sorted by confidence. To pick up the object, the gripper is first moved to a pre-grasp pose, positioned 10 centimeters away from the target along the gripper's z-axis. The gripper then approaches the target in a straight line while maintaining its orientation. Similarly, during placement, the robot first moves to a pre-place pose, followed by a final approach without altering the gripper orientation. The distance from the pre-place pose to the final placement pose is adjusted according to the object's size to avoid collisions during the transition to the pre-place position. With the waypoints and end-effector orientation constraints defined, we use cuRobo to generate the complete motion plan for robot pick and place while accounting for the object collision model. For each evaluation scene, we perform rejection sampling by generating 100 grasps using AnyGrasp and forming (grasp, placement pose) pairs to identify valid grasps for the specific placement pose predicted by our model that can be executed by the robot.

The pipeline itself can be used as a separate module. It makes no assumptions about the placement prediction method and is not specific in any way to other modules in our approach. This enables it to be used as a general system for evaluating placement pose prediction models. In \autoref{fig:example_placements} we show the system being used to perform insertion, hanging, and placement tasks. Combined with the synthetic dataset we create, this gives us a complete system for comparing different models and getting systematic results of success rate, mode coverage, and precision. Both motion planning and the pick-and-place simulation itself are GPU parallelizable, making the evaluation of new models even easier.

\subsection{AnyPlace Energy-based Model}
\label{appendix:energy}
Inspired by Implicit-PDF \cite{implicitpdf2021}, this model uses the same encoder as ours, but for the decoder, instead of explicitly predicting the placement pose, it includes two separate branches: one for placement location prediction and another for predicting the placement rotation energy. During training, we encourage all placement rotations in SO(3) that result in stable placements to have low energy by predicting the unnormalized log probability of the joint distribution between the latent state from the encoder and the SO3 rotation. This approach requires only minimal changes to the model. We use a negative log-likelihood loss for the rotation as shown below: 
\begin{equation}
\label{equation 3}
\mathcal{L}_{energy}(h, {R}_0) = -\log \, 
                        \frac{\exp(f(h, R_0) )}
                             {\sum_{i}^{N} \exp(f(h, R_i) )}
\end{equation}
During training, we estimate the normalization factor by sampling. In particular, we ran the model on 4096 randomly sampled SO3 rotations. The total loss is the sum of this loss and the same L1 loss for translation: 
\begin{equation}
    \mathcal{L}_{total}= \mathcal{L}_{translation} + \mathcal{L}_{energy}
\end{equation}
At test time, we can get the estimate of the full distribution over the rotations conditioned on the inputs to the model in the same way by randomly sampling thousands of rotations. The rotation with the lowest energy is then selected as the final placement rotation. 

We found that the translation loss consistently converges faster than the energy-based one during training as a result of its smaller scale. To address this imbalance, we experimented with various ways of weighting the two loss terms when computing the total loss, which led to slight performance improvements. In contrast, for the diffusion model, losses are computed at each diffusion step, and the scales of translation and rotation errors remain more balanced, resulting in more stable training. Overall, the energy-based model offers faster inference but lower accuracy compared to the diffusion model.

\section{Additional Result of Placement Precision}
\begin{figure*}[!h]
    \centering
\includegraphics[width=\textwidth]{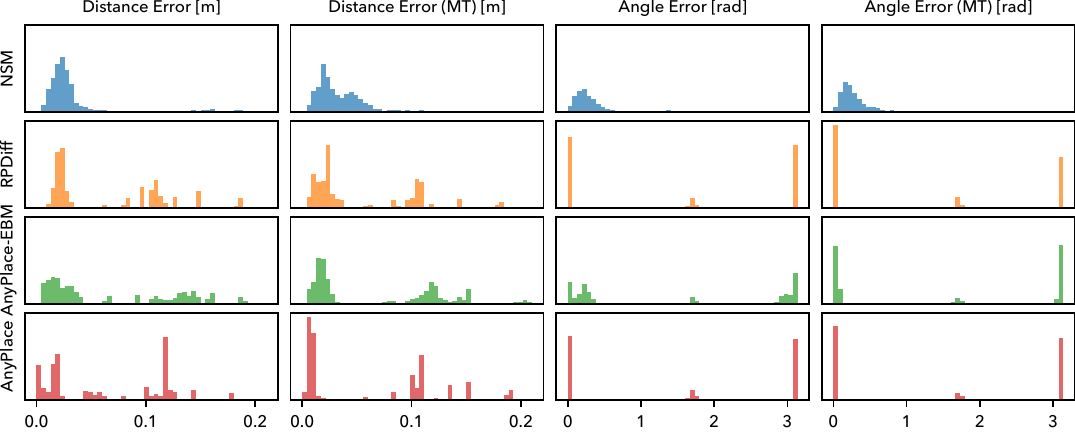}
    \caption{\textbf{Evaluation of Distance and Angle Errors in Insertion Tasks.} We plot histograms of the translation and rotation error in pose prediction for each approach trained on single-task data and multi-task data. We base the error on the position and rotation to the closest viable placement pose. The rotation error is in the range $[0, \pi]$. For both translation and rotation, we use 50 equally-sized bins covering the range of the variable to produce the histograms.
  }
    \label{fig:full_precision_plot}
\end{figure*}

In this section, we provide additional results on the placement precision of fine-grained insertion tasks for all models trained on single-task data. In \autoref{fig:full_precision_plot}, we show the distribution of distance errors for each approach on the insertion tasks. For both single-task and multi-task training, we observe that \ourmethod achieves smaller errors and does so more reliably than the baselines, in terms of translation and rotation prediction.
 
\subsection{Additional Analysis of Real Robot Experiments}

\begin{figure*}[!h]
    \centering
\includegraphics[width=\textwidth]{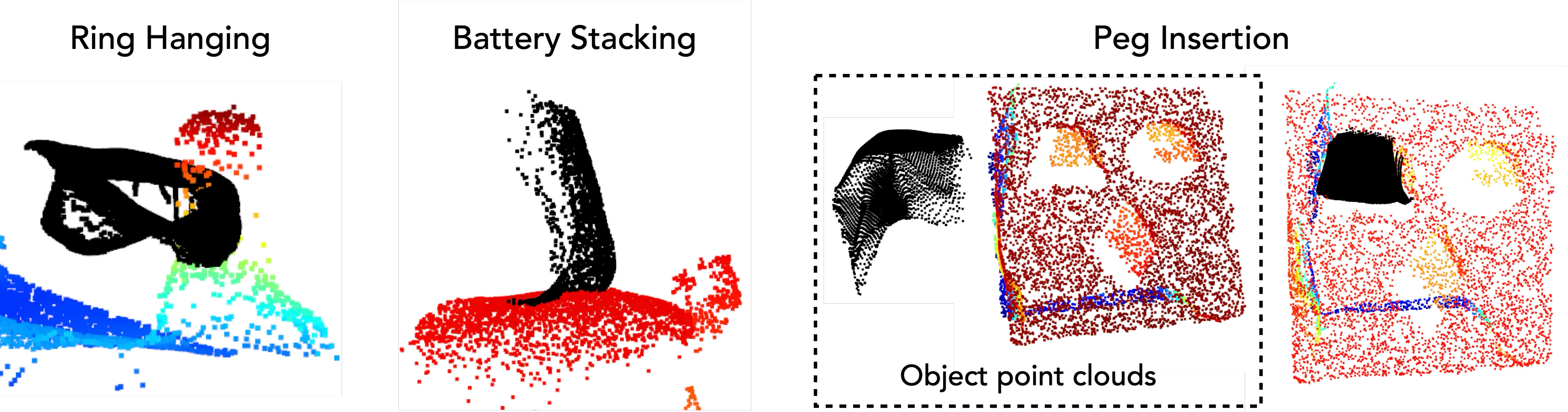}
   \caption{\textbf{Object point clouds at predicted placement poses in failure cases.} The point cloud of the object being placed is shown in black. }
    \label{fig:real_predict_fail}
\end{figure*}

\begin{figure*}[t]
    \centering
\includegraphics[width=0.8\textwidth]{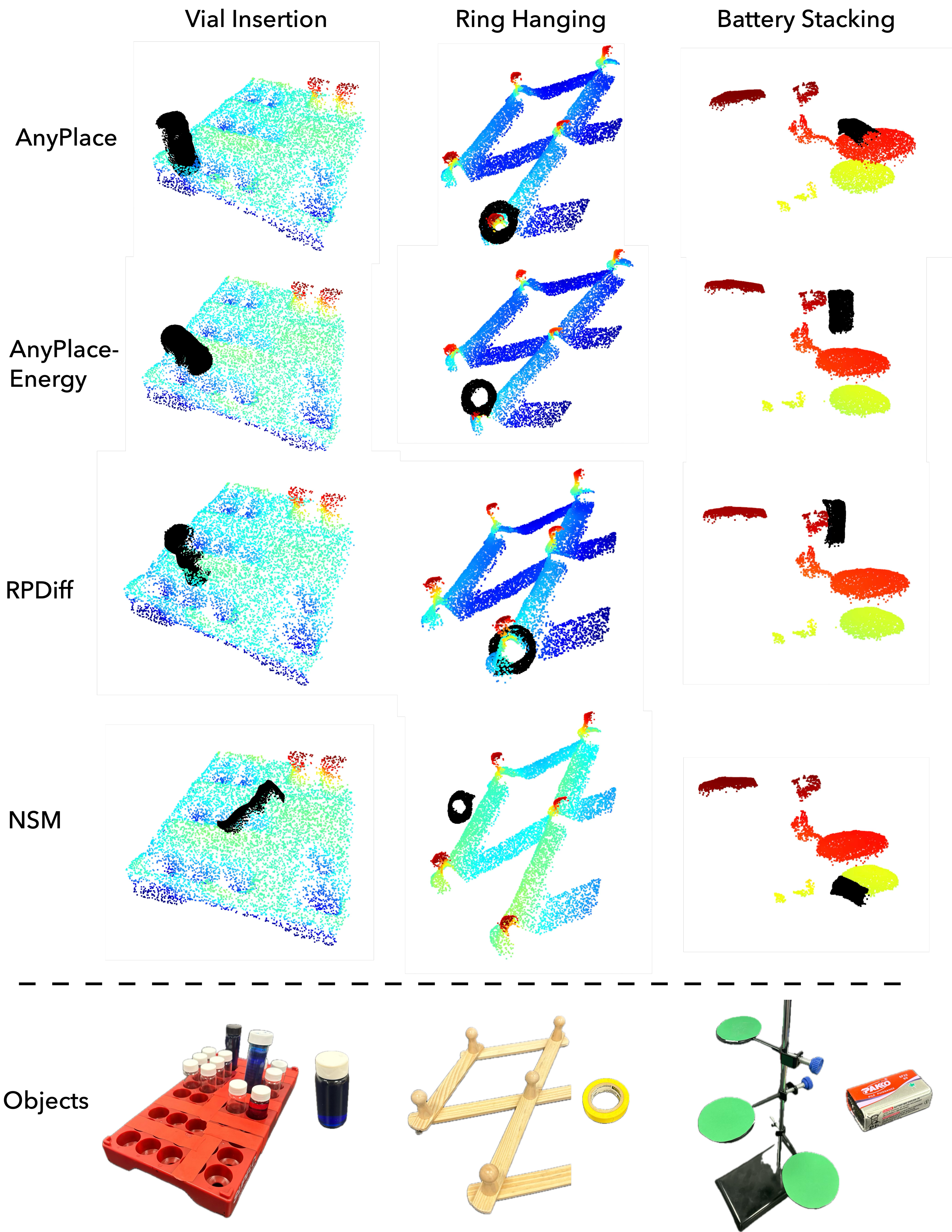}
    \caption{\textbf{Visualization of Object Point Clouds at Predicted Poses from Different Models.} 
Quantitative evaluations were conducted on three real-world tasks: vial insertion, ring hanging, and battery stacking. The figure shows examples of predicted placement poses visualized using point clouds. 
The point cloud of the placed object (target object) is shown in black.
  }
    \label{fig:real_quan}
\end{figure*}

We systematically evaluate the performance of all baseline models and AnyPlace across three tasks on a real robot: vial insertion, ring hanging, and battery stacking. As shown in the \autoref{fig:real_quan}, AnyPlace successfully predicts correct and precise placement poses using partial point clouds for all tasks, due to its ability to focus on the local placement region and the robustness of its diffusion-based pose prediction model. Although AnyPlace-Energy uses the same high-level placement location module, we observe that the energy-based model often predicts incorrect poses in all tasks, both in position and orientation. Similarly, RPDiff frequently converges to the same placement pose, failing to capture diverse placement locations. It also struggles significantly with predicting accurate positions. Due to the fixed cropping size introduced in RPDiff, its generalization to various objects is limited, causing inaccurate prediction. For example, in the vial insertion and ring hanging tasks, the predicted poses often cause the placing object's point cloud to deeply intersect with the base objects. For battery stacking, the model predicts a pose that is far from the correct placement location. Finally, NSM, which uses a simple regression decoder, is unable to predict any meaningful placement poses.

\begin{figure*}[!h]
    \centering
\includegraphics[width=0.9\textwidth]{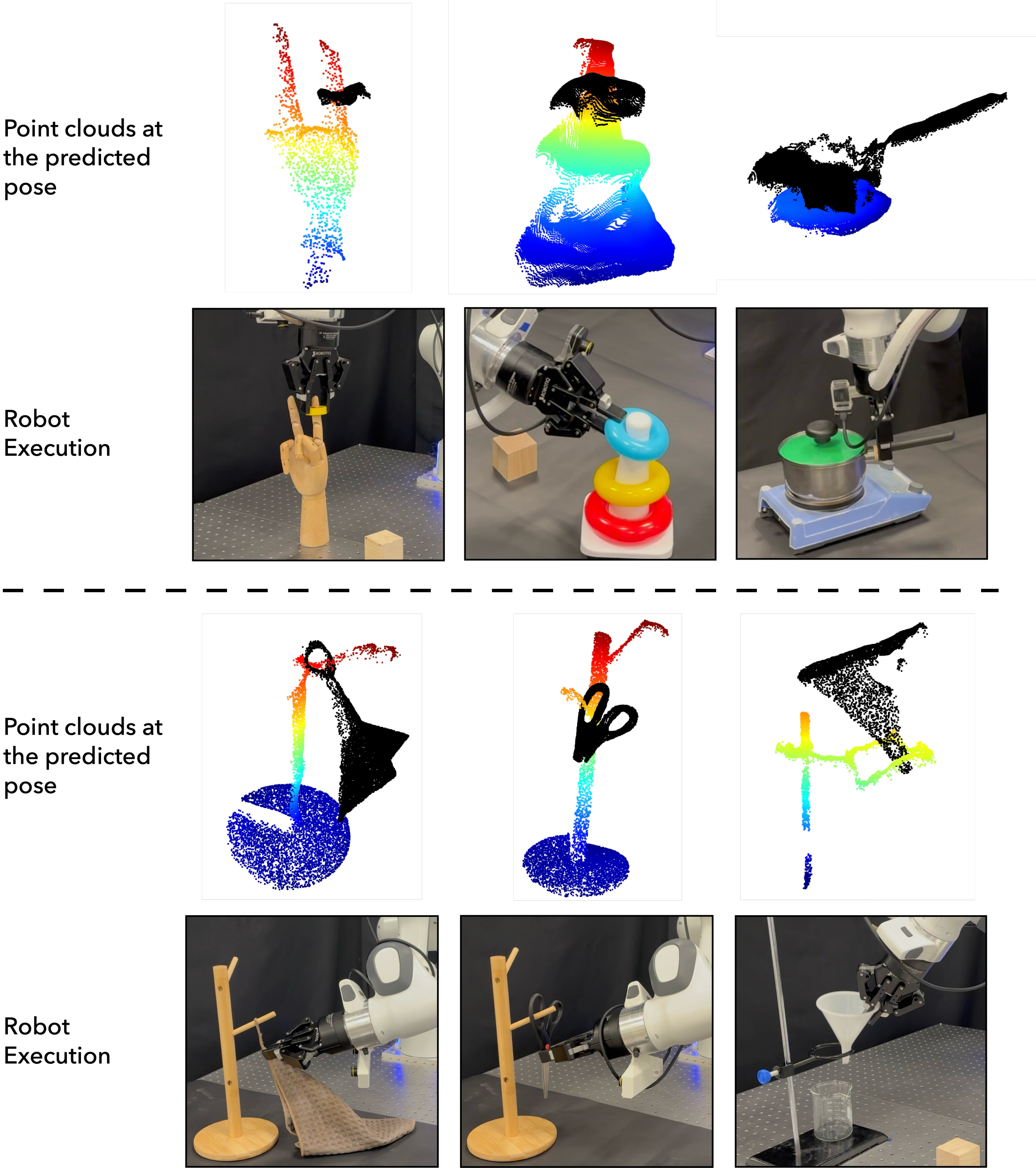}
   \caption{\textbf{Object point clouds at predicted placement poses in failure cases.} The point cloud of the object being placed is shown in black. }
    \label{fig:real_predict_additional}
\end{figure*}

To further evaluate the generalization and robustness of AnyPlace in the real world, we conducted extensive experiments across different placement scenarios. We visualize the point clouds at the predicted placement poses in \autoref{fig:real_predict_additional}. It is clear that AnyPlace is capable of handling unseen objects and noisy point clouds. For example, in many insertion tasks, the model accurately predicts placement poses even when dealing with complex, unseen objects, such as focusing on individual fingers when hanging a ring onto a hand. This level of generalization is achieved by focusing on local regions rather than relying on global information, which may contain redundant or irrelevant details. Moreover, training the low-level model on a diverse synthetic dataset enables it to capture general concepts and representations across different placement scenarios. For example, the model was never trained with a funnel or the ring clamp, yet it successfully predicts the correct placement poses for these novel objects.

\end{document}